\documentclass[10pt,journal,compsoc]{IEEEtran}

\ifCLASSOPTIONcompsoc
  \usepackage[nocompress]{cite}
\else
  \usepackage{cite}
\fi
\ifCLASSINFOpdf
\else
\fi
\usepackage{graphicx}
\usepackage{xcolor}
\usepackage{color}
\usepackage{mathrsfs}
\usepackage{amsmath}
\usepackage{amssymb}
\usepackage{ragged2e}
\usepackage{multirow}
\usepackage{multicol}
\usepackage{amssymb}
\usepackage{listings}
\usepackage{subcaption}
\usepackage{bbm}
\usepackage[breaklinks=true,bookmarks=true,citecolor=citecolor,colorlinks,pagebackref=true]{hyperref}
\usepackage{colortbl}
\usepackage{float}

\def\ie{\emph{i.e.}}
\def\eg{\emph{e.g.}}
\def\etal{{\em et al.}}
\def\etc{{\em etc.}} 
\definecolor{dkgreen}{rgb}{0,0.6,0}
\definecolor{gray}{rgb}{0.55,0.55,0.55}
\definecolor{mauve}{rgb}{0.58,0,0.82}
\definecolor{citecolor}{HTML}{0071bc}
\definecolor{shadecolor}{rgb}{0.94,0.94,0.94}

\newcommand{\graytext}[1]{\textcolor{gray}{#1}}
\newcommand{\bluetext}[1]{\textcolor{citecolor}{#1}}
\newcolumntype{g}{>{\columncolor{shadecolor}}c}

\usepackage{booktabs}
\usepackage{array, caption, threeparttable}
\usepackage[font=small,labelfont=bf]{caption}
\begin{document}
\title{UniMatch V2: Pushing the Limit of Semi-Supervised Semantic Segmentation}
\author{
Lihe Yang, Zhen Zhao$^\dag$, and Hengshuang Zhao$^\dag$
\IEEEcompsocitemizethanks{

\IEEEcompsocthanksitem Lihe Yang and Hengshuang Zhao are with The University of Hong Kong. \\
Email: lihe.yang.cs@gmail.com, hszhao@cs.hku.hk.

\IEEEcompsocthanksitem Zhen Zhao is with Shanghai AI Laboratory. \\
Email: zhaozhen@pjlab.org.cn

\IEEEcompsocthanksitem $^\dag$Corresponding authors: Zhen Zhao and Hengshuang Zhao.
}
}

\markboth{}
{Shell \MakeLowercase{\textit{et al.}}: Bare Demo of IEEEtran.cls for Computer Society Journals}
\IEEEtitleabstractindextext{

\begin{abstract}
    \justifying 
    Semi-supervised semantic segmentation (SSS) aims at learning rich visual knowledge from cheap unlabeled images to enhance semantic segmentation capability. Among recent works, UniMatch~\cite{unimatch} improves its precedents tremendously by amplifying the practice of weak-to-strong consistency regularization. Subsequent works typically follow similar pipelines and propose various delicate designs. Despite the achieved progress, strangely, even in this flourishing era of numerous powerful vision models, almost all SSS works are still sticking to 1) using outdated ResNet encoders with small-scale ImageNet-1K pre-training, and 2) evaluation on simple Pascal and Cityscapes datasets. In this work, we argue that, it is necessary to switch the baseline of SSS from ResNet-based encoders to more capable ViT-based encoders (\eg, DINOv2) that are pre-trained on massive data. A simple update on the encoder (even using 2$\times$ fewer parameters) can bring more significant improvement than careful method designs. Built on this competitive baseline, we present our upgraded and simplified UniMatch V2, inheriting the core spirit of weak-to-strong consistency from V1, but requiring less training cost and providing consistently better results. Additionally, witnessing the gradually saturated performance on Pascal and Cityscapes, we appeal that we should focus on more challenging benchmarks with complex taxonomy, such as ADE20K and COCO datasets. Code, models, and \emph{logs of all reported values}, are available at \texttt{\url{https://github.com/LiheYoung/UniMatch-V2}}.
\end{abstract}

\begin{IEEEkeywords}
    Semi-Supervised Learning, Semantic Segmentation, Weak-to-Strong Consistency, Vision Transformer.
\end{IEEEkeywords}
}

\maketitle

\IEEEdisplaynontitleabstractindextext

\IEEEpeerreviewmaketitle

\section{Introduction}

Semantic segmentation~\cite{fcn, deeplabv2, pspnet, maskformer} plays a fundamental role in scene understanding by providing pixel-level class predictions. However, substantial dense annotations are required to learn a capable semantic segmentation model. For example, it takes around 1.5 hours to label a single image on Cityscapes~\cite{cityscapes} with merely 19 classes. This limitation greatly hinders the deployment of advanced models in critical applications without sufficient annotations. Therefore, to alleviate the burden of human annotators and decrease annotation costs, semi-supervised semantic segmentation (SSS) is attracting increasing attention. SSS aims to train a model with a small portion of labeled images and take full advantage of more unlabeled images. One of the most representative works recently is Segment Anything~\cite{sam}. It designs a semi-supervised data engine to gradually expand from small-scale human labels to automatically produced large-scale pseudo labels. Such methodology is universal and can be applied to many scenarios~\cite{depth_anything}. In this work, we especially focus on the task of semi-supervised semantic segmentation, which has been widely studied in recent years, covering extensive fields like natural image understanding~\cite{mittal2019semi, cps, cisc, corrmatch}, medical image analysis~\cite{uamt, bcp, zhao2023alternate, chi2024adaptive}, remote sensing interpretation~\cite{wang2021ranpaste, bandara2022revisiting, yuan2024dynamically}, \etc.

\begin{figure}
    \centering
    \includegraphics[width=0.9\linewidth]{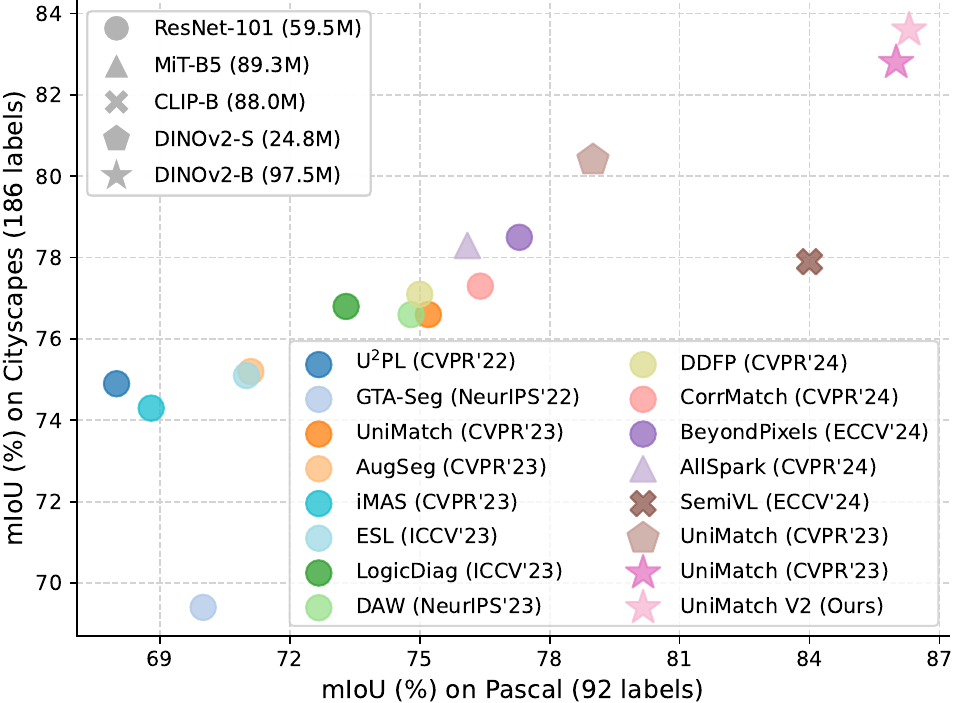}
    \caption{Performance of various methods across different pre-trained encoders \textbf{(upper-left legend)}. Under the ResNet-101 backbone, previous works struggle to further improve the best results. But with a simple update on the backbone (ResNet-101 $\rightarrow$ DINOv2-S $\rightarrow$ DINOv2-B) while keeping the method unchanged, the UniMatch~\cite{unimatch} performance is boosted significantly.}
    \label{fig:teaser}
    \vspace{-2mm}
\end{figure}

The core problem in SSS is how to effectively leverage unlabeled images. Existing works~\cite{cps, st++, unimatch} mostly follow the methodology of pseudo labeling (also called self-training)~\cite{pseudolabel, noisystudent}. The model first acquires initial semantic segmentation ability from labeled images, and then assigns pseudo labels (\ie, model predictions) to unlabeled images to expand available training samples. Such pseudo-labeling pipeline can be carried out either offline (\ie, multiple stages) or online (\ie, end-to-end). In an offline pipeline~\cite{st++}, the pseudo-labeling step is conducted only when the model has been sufficiently trained on labeled images. In contrast, for an online pipeline~\cite{pseudolabel}, the model predicts pseudo labels in each training iteration for the sampled unlabeled batch. From the very start of training, the model is jointly optimized on manually labeled images and pseudo-labeled images. Both offline and online roadmaps have witnessed great development in the past few years.

Dating back to three years ago, ST++~\cite{st++} demonstrates that, a plain offline self-training pipeline~\cite{pseudolabel} is indeed superior to previous online ones~\cite{cutmixseg, pseudoseg}, as long as injecting appropriate strong data augmentations to unlabeled images. Although such an offline strategy can ensure the quality of pseudo labels, it is not elegant enough, requiring three separate stages. In view of this, UniMatch~\cite{unimatch} revisits the weak-to-strong consistency regularization, which is simplified and popularized by FixMatch~\cite{fixmatch} originally in semi-supervised classification. As an elegant online self-training framework, FixMatch estimates pseudo labels on weakly-augmented (\eg, cropped) clean images and uses these labels to supervise the training of corresponding strongly-augmented (\eg, color-jittered) images. To select reliable pseudo labels to learn, it pre-defines a confidence threshold and excludes model predictions not satisfying this criterion. Despite its simplicity and being proposed five years ago, UniMatch shows that, if equipped with strong spatial augmentations (\ie, CutMix~\cite{cutmix}), FixMatch is still a highly competitive baseline in SSS. It significantly outperforms all previous delicately designed methods before 2023.

FixMatch~\cite{fixmatch} harnesses rich visual knowledge by training on challenging strongly-augmented unlabeled images. However, the strong augmentations are constrained in the input space, \ie, merely applying color and spatial distortions to raw images. This prohibits the model from pursuing invariant representations under a broader augmentation space. Thus, to further promote the spirit of weak-to-strong consistency in FixMatch, UniMatch~\cite{unimatch} employs an additional feature-level augmentation stream as a supplement to the input-level stream. It finds a simplest channel-wise Dropout~\cite{dropout} on intermediate features works the most effectively. Moreover, to fully explore the original input-level augmentation space, it designs a dual-stream augmentation strategy at the input level. Two strongly-augmented images are jointly sampled from their shared weakly-augmented version through a random data augmentation pool. They are passed into the model in two parallel streams for training. With the two key practices (feature-level augmentation and dual-stream augmentation), UniMatch further improves the FixMatch performance remarkably. Due to the great simplicity and easily reproduced strong results, many subsequent works in SSS build their framework on UniMatch~\cite{daw, beyond, shin2024revisiting, semivl} directly or on the more basic FixMatch~\cite{corrmatch, allspark} reproduced by UniMatch.

However, after checking recent works in SSS, we noticed that their methods are becoming increasingly sophisticated. More importantly, even with these carefully designed modules, performance is usually only boosted by nearly 0.5\% on datasets like Pascal and Cityscapes. We can expect that, if we continue going this way, future works in this field will have much more difficulty in improving the current state-of-the-art (SOTA) results. As a result, they will have trouble in publishing their works just due to being ``not SOTA''. This will greatly hinder new ideas or new frameworks to flourish. As a fundamental research topic, development in SSS can provide valuable insights and guidelines for real-world computer vision (CV) applications on how to utilize unlabeled data effectively~\cite{sam, sam2, depth_anything, depth_anything_v2}. Therefore, we believe it is urgent to re-explore new meaningful roadmaps for future research on SSS.

Reviewing the development of SSS, numerous works have been published by ``designing novel methods''. In earlier years, new methods can bring a remarkable improvement of more than 5\%~\cite{cps}. However, recently, we no longer observe such significant advances. Most works only improve their precedents very marginally (around 0.5\%). It indicates that we have almost arrived at the boundary of the modeling capability of current models or the potential upper bound of evaluated benchmarks. Meantime, jumping out of our narrow SSS field, in the past few years, the wider CV community has witnessed tremendous progress in 1) new model architectures, \eg, vision transformers~\cite{vit, swin, biformer}, better convolutional networks~\cite{convnext, repvgg}, 2) better pre-training strategies~\cite{clip, ibot, eva}, especially vision-alone self-supervised learning methods~\cite{moco, simclr, beit, mae, dinov2}, and 3) leveraging ultra-massive data (over 100M) for pre-training~\cite{clip, evaclip, openclip, dinov2}. Despite these exciting progress, pitifully, none of them have been well integrated into our SSS field. Recent SSS works are still sticking to the outdated ImageNet-1K~\cite{deng2009imagenet} pre-trained ResNet encoders~\cite{resnet}. To some extent, it can be understood because most previous SSS works have established the comparison baselines using the same encoder. It will be risky and costly to re-benchmark existing methods with new architectures. However, to make the SSS works have a broader impact, we believe it is worthy and urgent to do so, because there is no guarantee that insights obtained from these outdated encoders with small-scale pre-training can be safely transferred to modern architectures with large-scale pre-training.

Among latest SSS works, there are two exceptions discarding the ResNet encoders. One is SemiVL~\cite{semivl}, building on the CLIP-ViT-B model~\cite{clip}, and the other is AllSpark~\cite{allspark}, using the MiT-B5~\cite{segformer} pre-trained encoder. Despite taking a step further, their used encoders are not powerful enough. And their experiments are not thorough enough to cover all datasets and fine-tuning strategies.

In this work, we aim at \emph{re-building a comprehensive new benchmark} for semi-supervised semantic segmentation with the most capable pre-trained model DINOv2~\cite{dinov2}. Thanks to its large-scale curated data and advanced training strategies, DINOv2 exhibits superior performance in widespread scenarios, \eg, classification, dense matching. As shown in Figure~\ref{fig:teaser}, without bells and whistles, a simple update on the pre-trained encoder from ResNet-101 to DINOv2-S (even 2$\times$ fewer parameters) boosts the UniMatch~\cite{unimatch} performance by over 3\% on Pascal and 4\% on Cityscapes. And DINOv2-B is much stronger than other encoders of similar scales, such as CLIP-B~\cite{clip} and MiT-B5~\cite{segformer}. Impressed by the results, we re-conduct all core experiments of UniMatch V1~\cite{unimatch} and its baseline FixMatch~\cite{fixmatch} with DINOv2.

Built on this strong baseline (UniMatch + DINOv2), we further present our \emph{upgraded and simplified UniMatch V2}. It inherits the core spirit of weak-to-strong consistency regularization from V1. But differently, V2 uses fewer strongly-augmented streams for learning. It fuses the feature-level Dropout and input-level augmentations into a single stream. Moreover, to fully explore the joint augmentation space, we design a Complementary Dropout at the feature level. It decomposes the feature maps along the channel dimension into two disjoint and complementary sets. The two non-overlapping feature sets can be considered as two different yet meaningful views of an image. For example, one set of features may be sensitive to textures, while another set is responsible for the structure information. We then forward these two complementary features into the shared decoder for dual-stream learning. Compared with the dual-stream practice in V1 (twice random image augmentations), our Complementary Dropout can produce better dual views, which proves more effective in practice.

This work greatly updates our CVPR'23 work UniMatch V1~\cite{unimatch}, with \textbf{multiple new contributions}: \textbf{(1)} We re-evaluate UniMatch V1 as well as its baseline FixMatch~\cite{fixmatch} (two most widely adopted methods in SSS) with the most capable vision foundation model DINOv2~\cite{dinov2}. \textbf{(2)} We simplify V1 by fusing its feature-level and input-level augmentations into a single learnable stream. Additionally, we propose a Complementary Dropout to fully harness dual-stream training. With the two upgrades, V2 shows more efficient training and better performance than V1. \textbf{(3)} We conduct extensive experiments, including but not limited to i) four popular semantic segmentation datasets, ii) a wide range of vision encoders: ResNet~\cite{resnet}, DINOv2~\cite{dinov2}, SAM~\cite{sam}, MiT~\cite{segformer}, BEiT~\cite{beit}, iii) fine-tuning \emph{vs.} freezing the pre-trained encoder, iv) comprehensive ablation studies on different frameworks and hyper-parameters across abundant settings, v) real-world large-scale SSS setting with considerable labeled images and much more unlabeled images, and vi) broader semi-supervised scenarios, \eg, remote sensing changing detection and image classification.
\section{Related Work}

\subsection{Semi-Supervised Learning}

Semi-supervised learning (SSL)~\cite{zhu2005semi, chapelle2009semi, van2020survey} studies how to better utilize unlabeled images. It is a fundamental and long-standing problem in machine learning, with extensive applications ranging from classical image classification~\cite{pseudolabel} to modern CV and NLP foundation models~\cite{sam, depth_anything, huang2022large}. Since the rise of deep neural networks, it has earned increasing attention. The reason behind this is that deep models are capable of fitting various training samples and then generalizing, but they are extremely hungry for labeled data, which are usually costly to acquire. The lack of training samples will greatly limit the generalization ability of modern networks. Therefore, researchers resort to cheap unlabeled data that widely exist and are effortless to collect. They can play a valuable role in increasing the data coverage and enhancing the model transferring capability.

To effectively incorporate unlabeled data, there are two mainstream methods. One is called \emph{entropy minimization}~\cite{grandvalet2005semi, rosenberg2005semi, noisystudent, zoph2020rethinking, pham2021meta}, popularized by a straightforward self-training pipeline~\cite{pseudolabel}. Typically, it first trains a teacher model on initial labeled data. And then this teacher model predicts pseudo labels on additional unlabeled data. The predicted logits are usually sharpened or post-processed as one-hot pseudo labels in classification. These hard labels with minimized entropy can serve as supervision signals to re-train a student model on the unlabeled data. To better use this pipeline, one key insight from Noisy Student~\cite{noisystudent} is that we should inject strong data augmentations and model augmentations when re-training the student model to increase its learning difficulty.

Despite its effectiveness, the offline self-training pipeline has been rarely used, due to its inconvenient three stages. Many recent works prefer the end-to-end pseudo-labeling frameworks powered by \emph{consistency regularization}~\cite{mt, uda, mixmatch, remixmatch, fixmatch}. The core idea is to align the poisoned predictions with more accurate predictions. From the model aspect, Mean Teacher~\cite{mt} maintains an EMA teacher model to produce better pseudo labels for the student. From the data aspect, FixMatch~\cite{fixmatch} uses the high-quality prediction on a clean image to supervise the training of corresponding strongly-augmented hard image. Due to its simplicity and efficacy, numerous subsequent works~\cite{lassl, wang2022freematch, du2023semi, iomatch, gan2024erasing} build on it. Among them, FlexMatch~\cite{flexmatch} replaces the global confidence threshold with class-aware thresholds by considering the class-wise learning status. DST~\cite{chen2022debiased} decouples the prediction and learning of pseudo labels with separate heads to suppress model noise. Beyond semantic-level consistency, SimMatch~\cite{simmatch} applies an auxiliary regularization on instance-level relationships. To safely use all unlabeled data, ShrinkMatch~\cite{shrinkmatch} excludes confusion classes for uncertain samples to learn.

Our work inherits the weak-to-strong consistency regularization from FixMatch. Differently, we propose a dual-stream Complementary Dropout to craft better augmentation space. Besides, we focus on semi-supervised semantic segmentation, which is more laborious to acquire labeled data than the classification task of FixMatch.

\subsection{Semi-Supervised Semantic Segmentation}

Semi-supervised semantic segmentation (SSS) is an essential subfield of SSL, with extensive applications in scene understanding~\cite{mittal2019semi, cps, cisc, corrmatch}, medical image analysis~\cite{uamt, bcp, zhao2023alternate, chi2024adaptive}, and remote sensing interpretation~\cite{wang2021ranpaste, bandara2022revisiting, yuan2024dynamically}. Earlier works~\cite{mittal2019semi, souly2017semi} make some pioneering efforts in borrowing the methodology of GANs~\cite{gan} to SSS. They train the segmentation model as a generator to produce predictions that can fool the network used to discriminate pseudo labels from manual labels. Despite inspiring, such frameworks are very hard to train. Later works~\cite{cct, feng2020dmt, ke2020guided, mendel2020semi, pseudoseg, zhou2021c3, zhong2021pixel, alonso2021semi, zhang2021robust, cac, redistributing, guan2022unbiased, liu2022bootstrapping, kwon2022semi, zhang2022region} till now, mostly follow the progress of SSL, proposing simpler yet useful designs from the perspectives of \emph{entropy minimization} and \emph{consistency regularization}. These two principles are usually incorporated together.

During this trend, French \etal~\cite{cutmixseg} highlight the necessity of injecting Cutout~\cite{cutout} and CutMix~\cite{cutmix} into unlabeled images to challenge the model to learn. Then CPS~\cite{cps} reveals that CutMix can work more effectively with a co-training framework~\cite{blum1998combining, dmt, cotraining}. AEL~\cite{ael} further designs an adaptive CutMix to bias the sampled data towards under-performing classes. These three works use an online pseudo-labeling framework. Their pseudo label quality cannot be guaranteed at the early training stages. In contrast, ST++~\cite{st++} and SimpleBaseline~\cite{simplebaseline} adopt a three-stage self-training pipeline, pseudo labeling only when the model has been fully trained on labeled data. Despite promising results, they are not elegant enough.

To simplify the SSS methods and disclose what really matters, UniMatch~\cite{unimatch}, AugSeg~\cite{augseg}, and iMAS~\cite{imas} build simple yet effective frameworks based on the methodology of FixMatch~\cite{fixmatch}. They all emphasize the critical role of strong data augmentations on unlabeled images, such as color jittering and CutMix. They simply pre-define a confidence threshold to discard noisy pseudo labels, which proves very useful. In detailed designs, UniMatch~\cite{unimatch} reveals the benefits of dual-stream augmentations and feature-level augmentations. AugSeg~\cite{augseg} increases the randomness in RandAugment~\cite{randaug} for richer data augmentations. And iMAS~\cite{imas} employs adaptive augmentations and supervisions based on the model state.

Due to their simplicity and easily reproduced strong results, subsequent works mostly follow them, or the more basic FixMatch framework reproduced by UniMatch. Among the latest works, CorrMatch~\cite{corrmatch} leverages correlation maps to propagate and polish pseudo labels. AllSpark~\cite{allspark}, different from most works centered on unlabeled data, points out the value of labeled features. Most recently, SemiVL~\cite{semivl} combines a CLIP~\cite{clip} encoder into SSS via a spatial fine-tuning module and a language-aware decoder. Besides, BeyondPixels~\cite{beyond} finds it is helpful to additionally train a patch-level classifier for contexture information.

Despite the progress, these works do not adopt a powerful vision encoder, mostly sticking to the outdated ResNet encoder. We aim to re-benchmark all existing settings in SSS with the most capable DINOv2 encoder~\cite{dinov2}. Moreover, we present a simplified yet stronger UniMatch V2 framework based on this modern architecture.

\begin{figure*}[t]
    \centering
    \begin{subfigure}{0.135\textwidth}
        \centering
        \includegraphics[width=\textwidth]{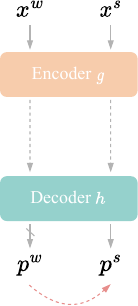}
        \caption{FixMatch~\cite{fixmatch}}
        \label{fig:fixmatch}
    \end{subfigure}
    \hfill
    \begin{subfigure}{0.161\textwidth}
        \centering
        \includegraphics[width=\textwidth]{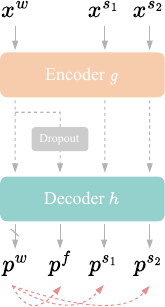}
        \caption{UniMatch V1~\cite{unimatch}}
        \label{fig:unimatch_v1}
    \end{subfigure}
    \hfill
    \begin{subfigure}{0.527\textwidth}
        \centering
        \includegraphics[width=\textwidth]{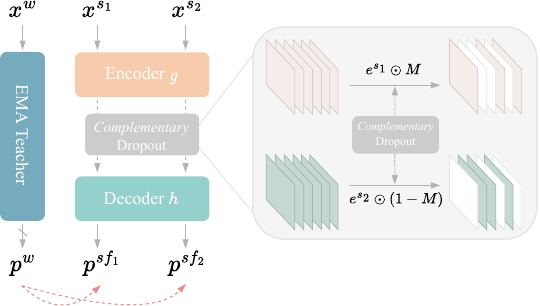}
        \caption{Our UniMatch V2, unified augmentations and complementary Dropout}
        \label{fig:unimatch_v2}
    \end{subfigure}
    \caption{Illustration of the evolution from FixMatch (a) to our prior UniMatch V1 (b), and to our current UniMatch V2 (c). FixMatch uses the prediction of a weakly-augmented image to supervise the corresponding strongly-augmented image. Based on FixMatch, UniMatch V1 brings a separate feature-level augmentation (\ie, Dropout) stream and an additional image-level augmentation stream. Our UniMatch V2, simpler and stronger than V1, unifies image-level and feature-level augmentations into a single stream (Section~\ref{sec:unimatch_v2_unify}) and presents complementary Dropout to craft better dual views (Section~\ref{sec:unimatch_v2_complementary}).}
    \label{fig:methods}
\end{figure*}

\section{Methodology}

Our proposed UniMatch V2 in this work, and our precedent work UniMatch V1~\cite{unimatch} are both based on FixMatch~\cite{fixmatch}. Thus we will primarily introduce this legacy framework in Section~\ref{sec:preliminaries}.
Then, we will go through our prior UniMatch V1 framework in Section~\ref{sec:unimatch_v1}, analyzing its advantages and limitations. Lastly, in Section~\ref{sec:unimatch_v2}, we will present our simplified UniMatch V2 framework, which is more efficient and more capable than V1.

\subsection{\label{sec:preliminaries}Preliminaries}

Generally, semi-supervised semantic segmentation (SSS) involves two datasets: a labeled dataset $\mathcal{D}^l = \{(x_i^l, y_i^l)\}$ and an unlabeled dataset $\mathcal{D}^u = \{x_i^u\}$, where $x_i$ is the $i$-th image and $y_i$ is its ground-truth semantic mask. In most cases, $\mathcal{D}^u$ is at a much larger scale than $\mathcal{D}^l$, \eg, 10$\times$ more images. The model learns initial visual knowledge from the small portion of labeled images, and then uses such knowledge to harvest the value of abundant unlabeled images.

To take full advantage of unlabeled data, FixMatch~\cite{fixmatch} (Figure~\ref{fig:fixmatch}) adopts a weak-to-strong consistency regularization framework~\cite{mixmatch, remixmatch}, which is trained in an end-to-end manner. Concretely, each mini-batch is composed of $B^l$ labeled images and $B^u$ unlabeled images. On the labeled images, the model is supervised by manually provided labels. This loss $\mathcal{L}^l$ can be formulated as:
\begin{equation}
    \mathcal{L}^{l} = \frac{1}{B^l}\sum_{i=1}^{B^l} \mathrm{H}(p^l_i, y^l_i),
\end{equation}
where $p^l_i$ is the model prediction on the $i$-th labeled image, and $y^l_i$ is its corresponding ground-truth mask. $\mathrm{H}$ is the widely used hard cross-entropy loss.

On the unlabeled images, the model first assigns pseudo labels (\ie, model predictions) to them and then learns in a self-teaching manner. Most importantly, in a weak-to-strong consistency regularization pipeline, the model predicts pseudo labels on the \emph{weakly-augmented image} $x^w$, but learns (\ie, trains) on its \emph{strongly-augmented version} $x^s$. The underlying logic of this asymmetric practice is that 1) the model produces higher-quality pseudo labels on the clean image $x^w$, but 2) directly training on $x^w$ will incur a minimal loss (little information), while $x^s$ is more appropriate for training since it can challenge the model to seek invariance under strong augmentations.

The $x^w$ is generated by feeding the original unlabeled image $x^w$ into a weak data augmentation pool $\mathcal{A}^w$, including basic image operators like cropping, resizing, and horizontal flipping. The $x^s$ is further yielded from $x^w$ through a strong data augmentation pool $\mathcal{A}^s$. $A^s$ consists of intensive color distortions and layout changes (\ie, CutMix~\cite{cutmix}). Formally, this data pre-processing follows:
\begin{equation}
    x^w = \mathcal{A}^w(x^u),~~~~x^s = \mathcal{A}^s(x^w).
\end{equation}

The model $f$ makes predictions on the two versions:
\begin{equation}
    p^w = f(x^w),~~~~p^{s} = f(x^s).
\end{equation}

As aforementioned, $p^w$ is considered as the pseudo label. In practice, the softmax output $p^w$ is further post-processed by an $\arg\max$ operator to become a hard one-hot label $\hat{p}^w$. It then supervises $p^s$ to train on the unlabeled data:
\begin{equation}
    \mathcal{L}^u = \frac{1}{B^u}\sum_{i=1}^{B^u} \mathbbm{1}(\max(p_i^w) \geq \tau)\mathrm{H}(p_i^s, \hat{p}_i^w),
\end{equation}
where $\mathbbm{1}(\max(p_i^w) \geq \tau)$ is a special design in FixMatch to alleviate the negative effect of noisy pseudo labels. It pre-defines a confidence threshold $\tau$ (\eg, 0.95). Pseudo labels not satisfying this threshold will be excluded from training. This mechanism ensures that, in early training iterations, the model is primarily optimized on high-quality manually labeled images, and then the training is progressively expanded to confidently pseudo-labeled images.

Finally, the joint loss for a mini-match is a simple interpolation of the labeled and unlabeled loss:
\begin{equation}
    \mathcal{L} = \mathcal{L}^l + \lambda\mathcal{L}^u,
\end{equation}
where $\lambda$ balances the effect of the unlabeled data. We simply set it as 1 in all our experiments.

\subsection{\label{sec:unimatch_v1}UniMatch V1: Unified Dual-Stream Augmentations}

Motivated by the impressive results of FixMatch when reproduced in SSS, UniMatch V1~\cite{unimatch} (Figure~\ref{fig:unimatch_v1}) aims at further strengthening its weak-to-strong consistency regularization from two perspectives, namely unified image-level and feature-level augmentations (Section~\ref{sec:uniperb}) and dual-stream augmentations (Section~\ref{sec:dusperb}).

\subsubsection{\label{sec:uniperb}Unified Augmentations of Image and Feature Levels}

FixMatch has achieved great success~\cite{liu2021unbiased, humbleteacher, softteacher, pixmatch, xiao2022learning, xu2022cross} by enforcing invariant model predictions under strong data augmentations $\mathcal{A}^s$. However, optimal $\mathcal{A}^s$ is not trivial to obtain, especially for images of specific domains, such as medical images and aerial images. It mostly requires domain experts to dig out the appropriate $\mathcal{A}^s$. This may limit a broader impact of our semi-supervised algorithms. More importantly, the augmentations (\eg, color jittering, CutMix) in FixMatch are all constrained in the image space, hindering the model from exploring more robust invariance under a broader augmentation space. Especially in this era of foundation models, some pre-trained encoders have been proven to be highly robust to color distortions~\cite{vaze2024no}. Thus, such image-only augmentations may be inadequate to fully unleash the potential of unlabeled images.

To this end, UniMatch V1 proposes to construct unified augmentations at both image and feature levels. Apart from the learnable stream of the strongly-augmented \emph{image} $x^s$, it maintains an additional stream of the strongly-augmented \emph{intermediate features}. Different from prior works~\cite{kuo2020featmatch, psmt} that combine all types of augmentations in a single stream, V1 reveals that it is better to disentangle different levels of augmentations into separate streams to avoid a single stream being excessively hard to learn. So V1 injects feature-level strong augmentations to weakly-augmented images $x^w$, rather than $x^s$. Formally, suppose a semantic segmentation model $f$ is composed of an encoder $g$ (\eg, ResNet~\cite{resnet}, DINOv2~\cite{dinov2}) and a decoder $h$ (\eg, ASPP~\cite{deeplabv2}, DPT~\cite{dpt}). Then, this process can be formulated as:
\begin{align}
e^w &= g(x^w),\\
p^{f} &= h(\mathcal{F}(e^w)),
\end{align}
where $e^w$ is the extracted clean intermediate features of the weakly-augmented image $x^w$. $\mathcal{F}$ is feature-level augmentations, such as Dropout~\cite{dropout}, adding uniform noise, or VAT~\cite{vat}. And $p^{f}$ is the decoder ($h$) output (\ie, model prediction) of the strongly-augmented features.

After incorporating this additional learnable stream on $p^f$, the overall unlabeled loss is updated as:
\begin{equation}
\small
    \mathcal{L}^u = \frac{1}{2B^u}\sum_{i=1}^{B^u} \mathbbm{1}(\max(p_i^w) \geq \tau)\big(\mathrm{H}(p_i^s, \hat{p}_i^w) + \mathrm{H}(p_i^f, \hat{p}_i^w)\big),
\end{equation}
where we assign equal loss weights for the image-level and feature-level learnable streams.

Through extensive ablation studies, V1 reveals a simple channel-wise Dropout (\texttt{nn.Dropout2d(0.5)} in PyTorch) works pretty well as the feature augmentation. It randomly selects half of the feature maps along the channel dimension and masks them out with zero value. We find that there is no need to adopt the computationally intensive VAT practice~\cite{vat, psmt} to perturb features. It is also worth noting that, in practice, a pre-trained encoder mostly outputs multi-level intermediate features for the input of the subsequent decoder. For example, the ASPP decoder~\cite{deeplabv2} takes both first-stage and final-stage features of the ResNet encoder. In such cases, we apply the feature augmentations to each feature volume independently.

\subsubsection{\label{sec:dusperb}Dual-Stream Augmentations}

The unified augmentations above successfully strengthen FixMatch by expanding the augmentation space. It enforces the model to seek robust representations under richer distortions, which is key to stronger generalization ability. Apart from this modification, UniMatch V1 further proposes to explore the original input-level augmentation space more thoroughly. To this end, it designs a dual-stream augmentation strategy. This strategy is motivated by the multi-view learning techniques in self-supervised learning and semi-supervised classification. For example, SwAV~\cite{caron2020unsupervised} uses a multi-crop technique to divide an image into multiple views of different resolutions, and then optimizes the local-to-global consistency among them. Similarly, ReMixMatch~\cite{remixmatch} produces multiple strongly-augmented images for the model to learn jointly.

Concretely, in our UniMatch V1, we first obtains two strongly-augmented images $(x^{s_1}, x^{s_2})$ from their shared weakly-augmented version $x^w$:
\begin{equation}
    x^{s_1} = \mathcal{A}^s(x^w),~~~~x^{s_2} = \mathcal{A}^s(x^w),
\end{equation}
where $x^{s_1}$ and $x^{s_2}$ are not equal, since the pre-defined strong data augmentation pool $\mathcal{A}^s$ are not deterministic. For example, the two versions may go through different types or strengths of color distortions. And their randomly selected CutMix regions may also be different.

We forward the two versions of images into the model in parellel. Their corresponding predictions $p^{s_1}$ and $p^{s_2}$ are jointly supervised by the high-quality prediction of their shared weakly-augmented version:
\begin{equation}
\small
    \mathcal{L}^u = \frac{1}{2B^u}\sum_{i=1}^{B^u} \mathbbm{1}(\max(p_i^w) \geq \tau)\big(H(p_i^{s_1}, \hat{p}_i^w) + H(p_i^{s_2}, \hat{p}_i^w)\big)
\end{equation}

We find in V1 that, despite the great simplicity, the dual-stream augmentations are highly beneficial to FixMatch. It is further validated that, the performance gain is non-trivial, not credited to a doubled unlabeled batch size. Compared with the single-stream augmentation, our design not only fully explores the original augmentation space, but also stabilizes the training. For instance, single-stream augmentation may produce excessively hard or easy images. In such cases, introducing an additional parallel stream for learning can serve as a valuable balance in model optimization (\ie, gradient descent). Moreover, we conjecture that supervising two strong views with a shared weak view can be conceptually regarded as enforcing consistency between these two strong views. Thus, our dual-stream practice shares the core spirits of contrastive learning~\cite{simclr, moco}, which is able to learn discriminative representations and has been proved to be highly meaningful to our SSS task~\cite{reco, u2pl}. We also adopt similar dual-stream methodology in our UniMatch V2, discussed next.

\begin{figure}
    \centering
    \hspace{-3mm}
    \includegraphics[width=0.92\linewidth]{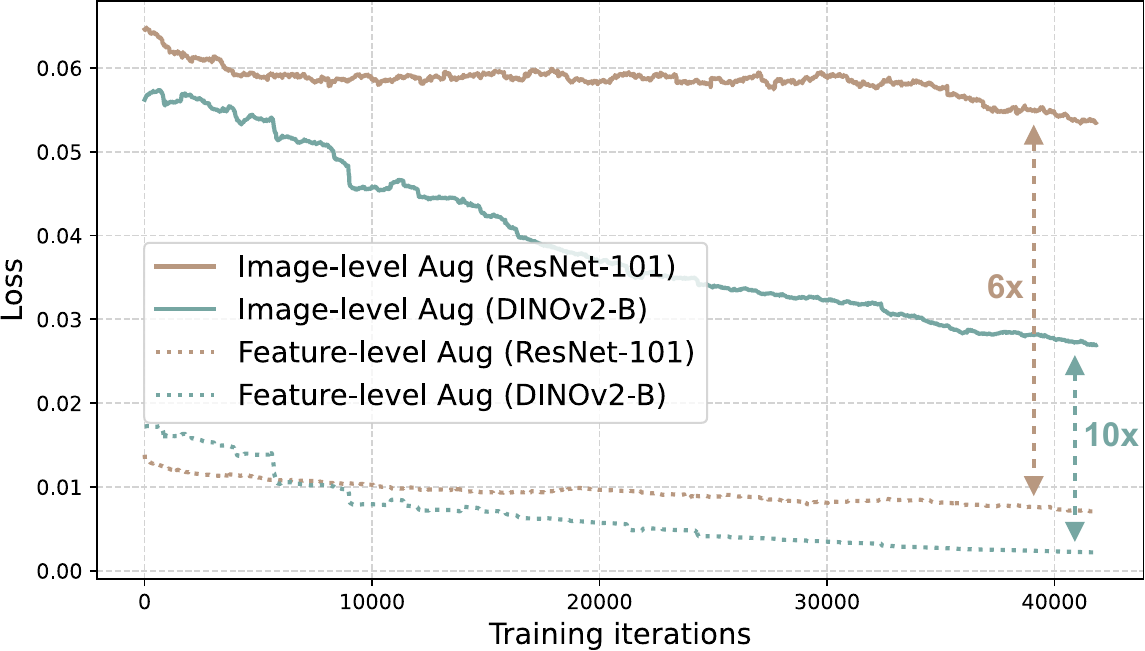}
    \caption{Comparison between the loss scales under image-level augmentations (\eg, color jittering, CutMix) and feature-level augmentation (\ie, Dropout). Loss incurred by image augmentations is much larger than that of feature augmentation.}
    \label{fig:loss_comparison}
\end{figure}

\subsubsection{Summary and Discussion}

Incorporating the above two designs of unified augmentations and dual-stream augmentations, there are three learnable streams in UniMatch V1, as well as one inference stream used to produce pseudo labels. The three streams enforce the model to keep consistent under multi-level random strong augmentations, greatly strengthening the weak-to-strong consistency regularization practice of FixMatch. The final loss on unlabeled data can be formulated as:
\begin{multline}
    \mathcal{L}^u_{v_1} = \frac{1}{2B_u}\sum_{i=1}^{B_u} \mathbbm{1}(\max(p_i^w) \geq \tau)\cdot \\
    \big(\mathrm{H}(p_i^f, \hat{p}_i^w) + \frac{1}{2}\big(\mathrm{H}(p_i^{s_1}, \hat{p}_i^w) + \mathrm{H}(p_i^{s_2}, \hat{p}_i^w)\big)\big).
\end{multline}
It can be observed that we evenly accumulate losses of the dual-stream augmentations. We also directly set equal weights for the two losses of input-level and feature-level augmentations. For the universality, we avoid carefully fine-tuning such hyper-parameters for better results. We still strictly obey such principles in our V2.

With these effective designs, UniMatch V1 successfully pushes the SSS results to a new bar. It has become the new baseline for many subsequent works~\cite{daw, beyond, shin2024revisiting, semivl}. However, it is still not optimal. For example, although there is only a single stream for final inference, three learnable streams are computationally heavy during training. Moreover, as exhibited in Figure~\ref{fig:loss_comparison}, the feature-level augmentation stream incurs a much smaller loss than the image-level augmentation, \eg, 6$\times$ smaller. When applying more advanced DINOv2 encoders, the feature-level loss will become even more negligible, contributing less to the final performance. Therefore, our UniMatch V2 aims to improve the training efficiency of V1, and further enhances its performance under capable vision foundation models.

\definecolor{codeblue}{rgb}{0.25,0.5,0.5}
\lstset{
  backgroundcolor=\color{white},
  basicstyle=\fontsize{7pt}{7pt}\ttfamily\selectfont,
  columns=fullflexible,
  breaklines=true,
  captionpos=t,
  commentstyle=\fontsize{7pt}{7pt}\color{codeblue},
  keywordstyle=\fontsize{7pt}{7pt},
}
\renewcommand{\lstlistingname}{Algorithm}

\captionsetup[lstlisting]{justification=raggedright, singlelinecheck=false}

\begin{figure}[t]
\vspace{-4.1mm}
\centering
\begin{lstlisting}[caption={Pseudocode of UniMatch V2 in a PyTorch style.}, label={alg:unimatch_v2}]
# f: network, composed of an encoder g and a decoder h
# f_ema: EMA teacher of f
# aug_w/aug_s: weak/strong image-level perturbations

# use binomial distribution to generate binary dropout masks
binomial = torch.distributions.binomial.Binomial(probs=0.5)

# cross-entropy loss function
criterion = torch.nn.CrossEntropyLoss()

for x_u in loader_u:
    # one weak view and two strong views as input
    x_w = aug_w(x_u)
    x_s1, x_s2 = aug_s(x_w), aug_s(x_w)
    
    # pseudo label obtained from weakly-augmented image
    pred_w = f_ema(x_w)
    mask_w = pred_w.argmax(dim=1).detach()
    
    # features (BxCxHxW) of dual strongly-augmented images
    feat_s = g(torch.cat((x_s1, x_s2)))
    
    # generate complementary channel-wise Dropout masks
    bs, dim = pred_w.shape[:2]
    mask_s1 = binomial.sample((bs, dim))
    mask_s2 = 1 - mask_s1
    mask = torch.cat((mask_s1, mask_s2)) * 2
    
    # perform Dropout on features
    feat_sf = feat_s * mask[..., None, None]
    
    # final decoder prediction after strong augmentations
    pred_sf = h(feat_sf)
    
    # loss from the dual streams
    loss_u = criterion(pred_sf, mask_w.repeat(2, 1, 1))
\end{lstlisting}
\vspace{-3mm}
\end{figure}

\subsection{\label{sec:unimatch_v2}UniMatch V2: \emph{Simpler} and \emph{Stronger}}

Primarily, from the architecture aspect, we construct our UniMatch V2 framework (Figure~\ref{fig:unimatch_v2}) on the most capable DINOv2 encoder. Almost all previous works, including our V1, still blindly use the outdated ResNet encoders, just to more conveniently compare with existing works. However, through our ablation studies (Table~\ref{tab:ablation_encoder}), the most lightweight DINOv2-Small encoder significantly outperforms the heaviest ResNet-152 encoder in our SSS task, using 3$\times$ fewer model parameters (24.8M \emph{vs.} 78.6M). Therefore, we appeal that it is urgent to switch to these modern encoders for a broader impact of future SSS works. And we will discuss our technical designs in the context of DINOv2.

Technically, in V2, we aim to 1) reduce the number of trainable streams in V1 for improved training efficiency, 2) still maintain the core spirit of unified image-level and feature-level augmentations, and 3) achieve further better results than V1 under the modern encoders.

\subsubsection{\label{sec:unimatch_v2_unify}Single-Stream Unified Augmentations}

As revealed in Figure~\ref{fig:loss_comparison}, the loss incurred by purely feature-level augmentations is very marginal compared with the loss under input-level augmentations. The gap is even larger when updating the weak ResNet encoder to the powerful DINOv2 encoder. Through our experiments, we find there is no performance loss when removing the feature-level augmentation stream under DINOv2, \eg, four out of five settings on ADE20K are even improved by removing it. However, we believe that enforcing the model to be resistant to multi-level augmentations is still beneficial. To this end, we propose to unify the input-level and feature-level augmentations into \emph{a single stream}. This practice has three advantages: 1) the model can still pursue robust representation in a broad augmentation space, 2) there is no need to maintain separate streams for different forms of augmentations, thus the training efficiency is improved, and 3) DINOv2 is much more capable than previous ResNet, so it is beneficial to \emph{stack} various augmentations to further challenge it to learn, rather than \emph{decoupling} them as V1. The single-stream loss is formulated as:
\begin{align}
    p^{sf} &= h(\mathcal{F}(g(x^s))),\\
    \mathcal{L}^u &= \frac{1}{B^u}\sum_{i=1}^{B^u} \mathbbm{1}(\max(p_i^w) \geq \tau)\mathrm{H}(p_i^{sf}, \hat{p}_i^w).
\end{align}

\subsubsection{\label{sec:unimatch_v2_complementary}Complementary Channel-Wise Dropout}

Greatly impressed by the promising results of dual-stream augmentations in V1, we aim to amplify this methodology in our V2. Based on the above single-stream unified augmentations, an intuitive dual-stream practice is to randomly produce two strongly-augmented images, and then apply a random Dropout to their respective features for decoding. Despite promising, such a practice fails to fully decouple the dual streams for learning. Therefore, we further propose a \emph{Complementary Channel-Wise Dropout} to acquire two disjoint and complementary sets of features for the dual streams. Concretely, given a feature map $e^{s_1} \in \mathbb{R}^{B\times C\times H\times W}$ from the $x^{s_1}$ input stream, we randomly generate a dropout mask $M$ of the same shape as $e^{s_1}$. $M$ is a binary mask sampled from the binomial distribution (probability is 0.5), with half of its channels ($C/2$) all set as 1, and others as 0. With $M$, we can perform complementary channel-wise Dropout on features $e^{s_1}$ and $e^{s_2}$ (obtained from $g(x^{s_1})$ and $g(x^{s_2})$):
\begin{align}
    e^{s_1} &\leftarrow e^{s_1} \odot M \times 2,\\
    e^{s_2} &\leftarrow e^{s_2} \odot (1 - M) \times 2,
\end{align}
where the $M$ and $1-M$ are two complementary dropout masks. And the last scaling factor ``2'' is to ensure the output expectation is of the same scale as normal features.

Note that $e^{s_1}$ and $e^{s_2}$ are extracted by a shared encoder. They share the same characteristics per channel. Therefore, the complementary Dropout masks $M$ and $(1 - M)$ will enable us to obtain two disjoint sets of features with disjoint meanings for the dual-stream learning. Moreover, although the Dropout is inserted at the intersection of the encoder and decoder, the gradient will be back-propagated to the encoder, making the entire model more robust.

Given the dual complementary streams of input-level and feature-level augmentations, the final unlabeled loss in UniMatch V2 is formulated as:
\begin{equation}
    p^{sf_1} = h(e^{s_1}),~~~~p^{sf_2} = h(e^{s_2}),
\end{equation}
\begin{equation}
    \mathcal{L}^u_{v_2} = \frac{1}{2B^u}\sum_{i=1}^{B^u} \mathbbm{1}(\max(p_i^w) \geq \tau)\big(\mathrm{H}(p_i^{sf_1}, \hat{p}_i^w) + \mathrm{H}(p_i^{sf_2}, \hat{p}_i^w)\big)
\end{equation}

Another minor modification on V1 is, we maintain an exponentially moving averaged teacher model~\cite{mt, psmt, augseg}, rather than using the student model itself, to produce stable and better pseudo labels. Teacher parameters $\theta_t$ are updated alongside the student parameters $\theta^s$ by:
\begin{equation}
    \theta^t \leftarrow \gamma \times \theta^t + (1 - \gamma) \times \theta^s,
\end{equation}
where $\gamma$ is dynamically set as $\min(1 - \frac{1}{\mathrm{iter} + 1}, 0.996)$.

A PyTorch-like pseudocode of our UniMatch V2 is provided in Algorithm~\ref{alg:unimatch_v2}. It is conceptually simple to implement yet meantime highly effective.
\section{Experiment}

\begin{table*}[t]
\setlength\tabcolsep{3.2mm}
    \centering
    \begin{tabular}{lrccccccc}
    \toprule
    
    \multirow{2}{*}{\textbf{Pascal SOTAs}} & \multirow{2}{*}{Venue} & \multirow{2}{*}{Encoder} & \multicolumn{5}{c}{Ratio and absolute number of labeled images} & \multirow{2}{*}{\graytext{\#Params}} \\

    \cmidrule(lr){4-8}
    
    & & & 1/16 (92) & 1/8 (183) & 1/4 (366) & 1/2 (732) & Full (1464) \\
    
    \midrule

    Labeled Only & - & RN-101 & 45.1 & 55.3 & 64.8 & 69.7 & 73.5 & \graytext{59.5M} \\

    ST++~\cite{st++} & CVPR'22 & RN-101 & 65.2 & 71.0 & 74.6 & 77.3 & 79.1 & \graytext{59.5M} \\
		
    U$^2$PL~\cite{u2pl} & CVPR'22 & RN-101 & 68.0 & 69.2 & 73.7 & 76.2 & 79.5 & \graytext{59.5M} \\
		
    PS-MT~\cite{psmt} & CVPR'22 & RN-101 & 65.8 & 69.6 & 76.6 & 78.4 & 80.0 & \graytext{59.5M} \\

    GTA-Seg~\cite{gtaseg} & NeurIPS'22 & RN-101 & 70.0 & 73.2 & 75.6 & 78.4 & 80.5 & \graytext{59.5M} \\
    
    PCR~\cite{pcr} & NeurIPS'22 & RN-101 & 70.1 & 74.7 & 77.2 & 78.5 & 80.7 & \graytext{59.5M} \\

    iMAS~\cite{imas} & CVPR'23 & RN-101 & 68.8 & 74.4 & 78.5 & 79.5 & 81.2 & \graytext{59.5M} \\
    
    AugSeg~\cite{augseg} & CVPR'23 & RN-101 & 71.1 & 75.5 & 78.8 & 80.3 & 81.4 & \graytext{59.5M} \\
    
    \textbf{UniMatch V1}~\cite{unimatch} & CVPR'23 & RN-101 & 75.2 & 77.2 & 78.8 & 79.9 & 81.2 & \graytext{59.5M} \\

    \textbf{UniMatch V1}~\cite{unimatch} & CVPR'23 & CLIP-B & 77.9 & 80.1 & 82.0 & 83.3 & 84.0 & \graytext{88.0M} \\
    
    Diverse CoT~\cite{diverse} & ICCV'23 & RN-101 & 75.7 & 77.7 &  80.1 & 80.9 & 82.0 & \graytext{59.5M} \\

    ESL~\cite{esl} & ICCV'23 & RN-101 & 71.0 & 74.0 & 78.1 & 79.5 & 81.8 & \graytext{59.5M} \\

    LogicDiag~\cite{logicdiag} & ICCV'23 & RN-101 & 73.3 & 76.7 & 77.9 & 79.4 & - & \graytext{59.5M} \\

    DAW~\cite{daw} & NeurIPS'23 & RN-101 & 74.8 & 77.4 & 79.5 & 80.6 & 81.5 & \graytext{59.5M} \\

    DDFP~\cite{ddfp} & CVPR'24 & RN-101 & 75.0 & 78.0 & 79.5 & 81.2 & 82.0 & \graytext{59.5M} \\
    
    CorrMatch~\cite{corrmatch} & CVPR'24 & RN-101 & 76.4	& 78.5 & 79.4 & 80.6 & 81.8 & \graytext{59.5M} \\

    AllSpark~\cite{allspark} & CVPR'24 & MiT-B5 & 76.1 & 78.4 & 79.8 & 80.8 & 82.1 & \graytext{89.3M} \\
    
    BeyondPixels~\cite{beyond} & ECCV'24 & RN-101 & 77.3 & 78.6 & 79.8 & 80.8 & 81.7 & \graytext{59.5M} \\
    
    SemiVL~\cite{semivl} & ECCV'24 & CLIP-B & 84.0 & 85.6 & 86.0 & 86.7 & 87.3 & \graytext{88.0M} \\

    \midrule

    \rowcolor{shadecolor}
    Labeled Only (85.0) & - &  & 67.0 & 75.6 & 81.8 & 83.7 & 85.6 & \\

    \rowcolor{shadecolor}
    \textbf{UniMatch V2} & Ours & \multirow{-2}{*}{DINOv2-S} & 79.0 & 85.5 & 85.9 & 86.7 & 87.8 & \multirow{-2}{*}{\graytext{\textbf{24.8M}}} \\
    
    \midrule
    
    \rowcolor{shadecolor}
    Labeled Only (86.5) & - &  & 76.9 & 82.1 & 85.3 & 87.2 & 88.3 & \\

    \rowcolor{shadecolor}
    \textbf{UniMatch V2} & Ours & \multirow{-2}{*}{DINOv2-B} & \textbf{86.3} & \textbf{87.9} & \textbf{88.9} & \textbf{90.0} & \textbf{90.8} & \multirow{-2}{*}{\graytext{97.5M}} \\
    
    \bottomrule
    
    \end{tabular}
    \caption{Comparison with state-of-the-art methods on \textbf{Pascal} high-quality set. The number (\eg, 85.0) next to ``Labeled Only'' denotes the fully-supervised result (1464 precisely labeled images + 9118 coarsely labeled images).}
    \label{tab:pascal_origin}
\end{table*}

In this paper, our primary goal is to thoroughly update the outdated ResNet encoders with the powerful DINOv2 encoders in semi-supervised semantic segmentation (SSS). Therefore, we conduct comprehensive experiments, covering all previously used datasets and protocols, as well as not yet explored but more challenging and practical datasets that we wish to promote. In addition to the achieved state-of-the-art (SOTA) results of our UniMatch V2, we also provide the results of our UniMatch V1, our baseline FixMatch, and the labeled-only results with DINOv2. Moreover, we compile extensive ablation studies across a wide range of settings to demonstrate the effectiveness of our method. Considering the limited computational resources of some academic groups and to better facilitate future research, we also re-benchmark all the results \emph{under frozen pre-trained encoders} (less GPU memory, less training time).

Except the learning rate, which is necessary to re-explore an optimal value for the new encoder, we do not intensively fine-tune any hyper-parameters. We even keep the learning rate identical for all our explored settings. Therefore, despite our much stronger results than previous works, we believe there is still much room to further improve.

\subsection{Datasets}

We evaluate our UniMatch V2 on four popular benchmarks in semantic segmentation, \ie, Pascal~\cite{pascal}, Cityscapes~\cite{cityscapes}, ADE20K~\cite{ade20k}, and COCO~\cite{coco}. The first two datasets are widely used in SSS, but the last two are rarely evaluated, due to complex taxonomies. However, considering the saturated results on Pascal and Cityscapes, we believe we should pay more attention to ADE20K and COCO.

\textbf{Pascal}~\cite{pascal} (2012 version) contains 1464 images with high-quality semantic masks and 9118 images with less precise masks, spanning 21 classes. In this work, we select labeled images from the high-quality set, and treat all other images as unlabeled ones.

\textbf{Cityscapes}~\cite{cityscapes} is a classical urban scene dataset with 2975 images of 1024$\times$2048 resolution, covering 19 classes. Although the class space is relatively small, the label quality is very high due to the high resolution, containing many thin objects, such as traffic lights.

\textbf{ADE20K}~\cite{ade20k} is a rather challenging dataset composed of 150 classes. It includes 20210 labeled images. It is widely used in fully-supervised semantic segmentation, but rarely mentioned in semi-supervised setting. We suspect it is because previous ResNet-based models are too poor to achieve promising results on it. As we upgrade the encoder to DINOv2, we believe the semi-supervised results on ADE20K will be greatly improved. We hope it can serve as the main evaluation benchmark for future works.

\textbf{COCO}~\cite{coco} is a large and complex dataset. It is especially popular in object detection. In our semantic segmentation task, we use its 2017 version, containing 118287 labeled images and 81 classes. There are some pioneering attempts on this dataset in SSS, including our UniMatch V1.

Following previous practices, the labeled dataset is sampled from the entire dataset as a subset, and the remaining images are treated as unlabeled images.

\begin{table*}[t]
\setlength\tabcolsep{4.28mm}
    \centering
    \begin{tabular}{lrcccccc}
    \toprule

    \multirow{2}{*}{\textbf{Cityscapes SOTAs}} & \multirow{2}{*}{Venue} & \multirow{2}{*}{Encoder} & \multicolumn{4}{c}{Ratio and absolute number of labeled images} & \multirow{2}{*}{\graytext{\#Params}} \\

    \cmidrule(lr){4-7}
    
    & & & 1/16 (186) & 1/8 (372) & 1/4 (744) & 1/2 (1488) \\
    
    \midrule

    Labeled Only & - & RN-101 & 66.3 & 72.8 & 75.0 & 78.0 & \graytext{59.5M} \\
    
    U$^2$PL~\cite{u2pl} & CVPR'22 & RN-101 & 74.9 & 76.5 & 78.5 & 79.1 & \graytext{59.5M} \\
		
    PS-MT~\cite{psmt} & CVPR'22 & RN-101 & - & 76.9 & 77.6 & 79.1 & \graytext{59.5M} \\

    GTA-Seg~\cite{gtaseg} & NeurIPS'22 & RN-101 & 69.4 & 72.0 & 76.1 & - & \graytext{59.5M} \\
    
    PCR~\cite{pcr} & NeurIPS'22 & RN-101 & 73.4 & 76.3 & 78.4 & 79.1 & \graytext{59.5M} \\

    iMAS~\cite{imas} & CVPR'23 & RN-101 & 74.3 & 77.4 & 78.1 & 79.3 & \graytext{59.5M} \\
    
    AugSeg~\cite{augseg} & CVPR'23 & RN-101 & 75.2 & 77.8 & 79.6 & 80.4 & \graytext{59.5M} \\
    
    \textbf{UniMatch V1}~\cite{unimatch} & CVPR'23 & RN-101 & 76.6 & 77.9 & 79.2 & 79.5 & \graytext{59.5M} \\

    \textbf{UniMatch V1}~\cite{unimatch} & CVPR'23 & CLIP-B & 76.6 & 78.2 & 79.1 & 79.6 & \graytext{88.0M} \\

    Diverse CoT~\cite{diverse} & ICCV'23 & RN-101 & 75.7 & 77.4 & 78.5 & - & \graytext{59.5M} \\

    ESL~\cite{esl} & ICCV'23 & RN-101 & 75.1 & 77.2 & 78.9 & 80.5 & \graytext{59.5M} \\
    
    LogicDiag~\cite{logicdiag} & ICCV'23 & RN-101 & 76.8 & 78.9 & 80.2 & 81.0 & \graytext{59.5M} \\

    DAW~\cite{daw} & NeurIPS'23 & RN-101 & 76.6 & 78.4 & 79.8 & 80.6 & \graytext{59.5M} \\ 
    
    DDFP~\cite{ddfp} & CVPR'24 & RN-101 & 77.1 & 78.2 & 79.9 & 80.8 & \graytext{59.5M} \\
    
    CorrMatch~\cite{corrmatch} & CVPR'24 & RN-101 & 77.3 & 78.5 & 79.4 & 80.4 & \graytext{59.5M} \\

    AllSpark~\cite{allspark} & CVPR'24 & MiT-B5 & 78.3 & 79.2 & 80.6 & 81.4 & \graytext{89.3M} \\

    BeyondPixels~\cite{beyond} & ECCV'24 & RN-101 & 78.5 & 79.2 & 80.9 & 81.3 & \graytext{59.5M} \\
    
    SemiVL~\cite{semivl} & ECCV'24 & CLIP-B & 77.9 & 79.4 & 80.3 & 80.6 & \graytext{88.0M} \\

    \midrule

    \rowcolor{shadecolor}
    Labeled Only (83.8) & - &  & 77.2 & 80.2 & 81.7 & 82.4 &  \\

    \rowcolor{shadecolor}
    \textbf{UniMatch V2} & Ours & \multirow{-2}{*}{DINOv2-S} & 80.6 & 81.9 & 82.4 & 82.6 & \multirow{-2}{*}{\graytext{\textbf{24.8M}}} \\

    \midrule

    \rowcolor{shadecolor}
    Labeled Only (85.2) & - &  & 80.8 & 82.7 & 84.0 & 84.4 & \\

    \rowcolor{shadecolor}
    \textbf{UniMatch V2} & Ours & \multirow{-2}{*}{DINOv2-B} & \textbf{83.6} & \textbf{84.3} & \textbf{84.5} & \textbf{85.1} & \multirow{-2}{*}{\graytext{97.5M}} \\
    
    \bottomrule
    
    \end{tabular}
    \caption{Comparison with state-of-the-art methods on \textbf{Cityscapes}. The number (\eg, 83.8) next to ``Labeled Only'' denotes the fully-supervised result (2975 labeled images).}
    \label{tab:cityscapes}
\end{table*}

\begin{table*}[t]
\setlength\tabcolsep{2.85mm}
    \centering
    \begin{tabular}{lrccccccc}
    \toprule

    \multirow{2}{*}{\textbf{ADE20K SOTAs}} & \multirow{2}{*}{Venue} & \multirow{2}{*}{Encoder} & \multicolumn{5}{c}{Ratio and absolute number of labeled images} & \multirow{2}{*}{\graytext{\#Params}} \\

    \cmidrule(lr){4-8}

    & & & 1/64 (316) & 1/32 (631) & 1/16 (1263) & 1/8 (2526) & 1/4 (5052) \\
    
    \midrule

    CutMix~\cite{cutmixseg} & BMVC'20 & RN-101 & - & 26.2 & 29.8 & 35.6 & - & \graytext{59.5M} \\

    AEL~\cite{ael} & NeurIPS'21 & RN-101 & - & 28.4 & 33.2 & 38.0 & - & \graytext{59.5M} \\

    \textbf{UniMatch V1}~\cite{unimatch} & CVPR'23 & RN-101 & 21.6 & 28.1 & 31.5 & 34.6 & - & \graytext{59.5M} \\

    \textbf{UniMatch V1}~\cite{unimatch} & CVPR'23 & CLIP-B & 25.3 & 31.2 & 34.4 & 38.0 & - & \graytext{88.0M} \\

    SemiVL~\cite{semivl} & ECCV'24 & CLIP-B & 33.7 & 35.1 & 37.2 & 39.4 & - & \graytext{88.0M} \\

    \midrule

    \rowcolor{shadecolor}
    Labeled Only (49.0) & - &  & 26.1 & 32.7 & 37.1 & 39.8 & 42.7 & \\

    \rowcolor{shadecolor}
    \textbf{UniMatch V2} & Ours & \multirow{-2}{*}{DINOv2-S} & 31.5 & 38.1 & 40.7 & 44.4 & 45.8 & \multirow{-2}{*}{\graytext{\textbf{24.8M}}} \\

    \midrule

    \rowcolor{shadecolor}
    Labeled Only (54.1) & - & & 32.1 & 39.3 & 42.8 & 46.4 & 49.0 & \\

    \rowcolor{shadecolor}
    \textbf{UniMatch V2} & Ours & \multirow{-2}{*}{DINOv2-B} & \textbf{38.7} & \textbf{45.0} & \textbf{46.7} & \textbf{49.8} & \textbf{52.0} & \multirow{-2}{*}{\graytext{97.5M}} \\
    
    \bottomrule
    
    \end{tabular}
    \caption{Comparison with state-of-the-art methods on \textbf{ADE20K}. The number (\eg, 49.0) next to ``Labeled Only'' denotes the fully-supervised result (20210 labeled images).}
    \label{tab:ade20k}
\end{table*}

\subsection{Implementation Details}

We use the simple DPT~\cite{dpt} as our semantic segmentation model, built on DINOv2~\cite{dinov2}. We mainly report the results under DINOv2-Small and DINOv2-Base. We apply the Complementary Dropout with a probability 0.5. We adopt the same data augmentations as UniMatch V1~\cite{unimatch}. Specifically, weak augmentations $\mathcal{A}^w$ include random resizing between 0.5-2.0, random cropping, and horizontal flipping with probability 0.5. Strong augmentations $\mathcal{A}^s$ contain color jittering, grayscaling, gaussian blurring, and CutMix~\cite{cutmix}. Since the patch size of DINOv2 is 16, the training resolution has to be a multiplier of 14. For Pascal, ADE20K, and COCO, we use the training (\ie, cropped) size of 518, while for Cityscapes, the size is set as 798.

On Pascal and COCO, a mini-batch is evenly composed of 16 labeled images and 16 unlabeled images, while on Cityscapes and ADE20K, there are 8+8 labeled+unlabeled images. Unless otherwise specified, we conduct all experiments with four A100 GPUs.

We use the AdamW optimizer with weight decay of 0.01 for training. For the most important hyper-parameter learning rate (LR), we carefully seek the optimal value for the new DINOv2 encoder. Finally, for all datasets, we set the LR of the pre-trained encoder as 5e-6, and set the LR of the randomly initialized decoder as $40\times$ larger (\ie, 2e-4). We adopt a poly scheduler to decay the initial learning rate: $\mathrm{lr} \leftarrow \mathrm{lr} \times (1 - \frac{\mathrm{iter}}{\mathrm{total\_iter}})^{0.9}$. The model is trained for 60, 180, 60, and 20 epochs on Pascal, Cityscapes, ADE20K, and COCO, respectively. On Cityscapes, same as previous works, we adopt an online hard example mining (OHEM) loss~\cite{ohem} for labeled images, while in other cases, we adopt the standard cross-entropy loss. The confident threshold $\tau$ is set as 0.95 in all our experiments.

During inference, we only slightly interpolate the images to ensure their height and width are divisible by 14. On Cityscapes, following previous practice, we perform sliding window evaluation of the window size 798. We report the mean Intersection-over-Union (mIoU) metric ($\uparrow$) of the EMA teacher model. In most cases, it is approximately 0.1\% - 0.2\% better than the student after full convergence.

\begin{table*}[t]
\setlength\tabcolsep{2.6mm}
    \centering
    \begin{tabular}{lrccccccc}
    \toprule

    \multirow{2}{*}{\textbf{COCO SOTAs}} & \multirow{2}{*}{Venue} & \multirow{2}{*}{Encoder} & \multicolumn{5}{c}{Ratio and absolute number of labeled images} & \multirow{2}{*}{\graytext{\#Params}} \\

    \cmidrule(lr){4-8}

    & & & 1/512 (232) & 1/256 (463) & 1/128 (925) & 1/64 (1849) & 1/32 (3697) \\
    
    \midrule

    Labeled Only & - & XC-65 & 22.9 & 28.0 & 33.6 & 37.8 & 42.2 & \graytext{54.7M} \\

    PseudoSeg~\cite{pseudoseg} & ICLR'21 & XC-65 & 29.8 & 37.1 & 39.1 & 41.8 & 43.6 & \graytext{54.7M} \\

    PC$^2$Seg~\cite{pc2seg} & ICCV'21 & XC-65 & 29.9 & 37.5 & 40.1 & 43.7 & 46.1 & \graytext{54.7M} \\
    
    \textbf{UniMatch V1}~\cite{unimatch} & CVPR'23 & XC-65 & 31.9 & 38.9 & 44.4 & 48.2 & 49.8 & \graytext{54.7M} \\

    \textbf{UniMatch V1}~\cite{unimatch} & CVPR'23 & CLIP-B & 36.6 & 44.1 & 49.1 & 53.5 & 55.0 & \graytext{88.0M} \\

    CISC-R~\cite{cisc} & TPAMI'23 & XC-65 & 32.1 & 40.2 & 42.3 & - & - & \graytext{54.7M} \\
    
    LogicDiag~\cite{logicdiag} & ICCV'23 & XC-65 & 33.1 & 40.3 & 45.4 & 48.8 & 50.5 & \graytext{54.7M} \\

     AllSpark~\cite{allspark} & CVPR'24 & MiT-B5 & 34.1 & 41.7 & 45.5 & 49.6 & - & \graytext{89.3M} \\
    
    SemiVL~\cite{semivl} & ECCV'24 & CLIP-B & \textbf{50.1} & 52.8 & 53.6 & 55.4 & 56.5 & \graytext{88.0M} \\
    
    \midrule

    \rowcolor{shadecolor}
    Labeled Only (62.5) & - &  & 29.4 & 35.6 & 44.6 & 49.2 & 52.0 & \\

    \rowcolor{shadecolor}
    \textbf{UniMatch V2} & Ours & \multirow{-2}{*}{DINOv2-S} & 39.3 & 45.4 & 53.2 & 55.0 & 57.0 & \multirow{-2}{*}{\graytext{\textbf{24.8M}}} \\

    \midrule

    \rowcolor{shadecolor}
    Labeled Only (66.4) & - &  & 36.8 & 45.8 & 52.1 & 56.2 & 59.5 & \\

    \rowcolor{shadecolor}
    \textbf{UniMatch V2} & Ours & \multirow{-2}{*}{DINOv2-B} & 47.9 & \textbf{55.8} & \textbf{58.7} & \textbf{60.4} & \textbf{63.3} & \multirow{-2}{*}{\graytext{97.5M}} \\
    
    \bottomrule
    
    \end{tabular}
    \caption{Comparison with state-of-the-art methods on \textbf{COCO}. The number (\eg, 62.5) next to ``Labeled Only'' denotes the fully-supervised result (118287 labeled images).}
    \label{tab:coco}
\end{table*}

\begin{table*}[t]
\setlength\tabcolsep{2.25mm}
    \centering
    \begin{tabular}{lccccccccccccc}
    \toprule
    
    \multirow{3}{*}{\textbf{Ablation (DINOv2-B)}} & \multicolumn{5}{c}{Pascal} & \multicolumn{5}{c}{ADE20K} & \multicolumn{3}{c}{COCO} \\
    
    \cmidrule(lr){2-6}\cmidrule(lr){7-11}\cmidrule(lr){12-14}
    
    & 1/16 & 1/8 & 1/4 & 1/2 & Full & 1/64 & 1/32 & 1/16 & 1/8 & 1/4 & 1/512 & 1/256 & 1/128 \\

    & (92) & (183) & (366) & (732) & (1464) & (316) & (631) & (1263) & (2526) & (5052) & (232) & (463) & (925) \\

    \midrule
    
    Labeled Only & 76.9 & 82.1 & 85.3 & 87.2 & 88.3 & 32.1 & 39.3 & 42.8 & 46.4 & 49.0 & 36.8 & 45.8 & 52.1 \\
    
    FixMatch~\cite{fixmatch}$^\dag$ & 83.8 & 87.4 & 88.2 & 89.9 & 90.2 & 37.8 & 43.2 & 46.4 & 49.5 & 51.0 & 46.7 & 53.7 & 57.4 \\
    
    UniMatch V1~\cite{unimatch}$^\dag$ & 86.0 & 87.5 & 88.7 & \textbf{90.0} & 90.4 & 36.7 & 42.8 & 45.9 & \textbf{49.9} & 51.1 & 45.5 & 54.3 & 57.7 \\

    \rowcolor{shadecolor}
    UniMatch V2 & \textbf{86.3} & \textbf{87.9} & \textbf{88.9} & \textbf{90.0} & \textbf{90.8} & \textbf{38.7} & \textbf{45.0} & \textbf{46.7} & 49.8 & \textbf{52.0} & \textbf{47.9} & \textbf{55.8} & \textbf{58.7} \\
    
    \bottomrule
    
    \end{tabular}
    \caption{Ablation study against baseline methods (Labeled Only, FixMatch) and UniMatch V1 on Pascal, ADE20K and COCO. $\dag$: Different from their original implementations, we use an EMA teacher to produce stable and better pseudo labels.}
    \label{tab:ablation_baseline}
\end{table*}

\subsection{Comparison with State-of-the-Art Methods}

First, we want to emphasize our comparisons with previous state-of-the-art works are unfair, due to different encoders. We use much stronger encoders than other works, which is one of the main motivations of this work. Through the unfair comparisons, we hope to clearly reveal the superiority of DINOv2 in our SSS tasks, and appeal to more works to shift to this modern encoder. We will present \emph{fair ablation studies} under the same encoder in the next section.

\textbf{Pascal:} As shown in Table~\ref{tab:pascal_origin}, our DINOv2-S-based UniMatch V2 framework significantly outperforms previous ResNet-101-based frameworks. For example, under the setting of 1/8 (183) labeled images, the SOTA result with ResNet-101 is 78.6\% (reported in latest ECCV'24), while our result with DINOv2-S is 85.5\% (+6.9\%), even with over 2$\times$ fewer model parameters (59.5M \emph{vs.} 24.8M). Such a remarkable gain has never been achieved previously by modifying the frameworks. Moreover, compared with the latest works SemiVL~\cite{semivl} (based on CLIP~\cite{clip}) and AllSpark~\cite{allspark} (based on the MiT-B5~\cite{segformer}), our advantages are still huge. E.g., under the 1/4 (366) setting, our DINOv2-B result surpasses AllSpark and SemiVL by 9.1\% (79.8\% $\rightarrow$ 88.9\%) and 2.9\% (86.0\% $\rightarrow$ 88.9\%), respectively. We also report the ``Labeled only'' results, where only labeled images are used for training. We can observe that the labeled-only performance of DINOv2-S is even much better than the semi-supervised results of ResNet-101 in most cases. All these comparisons clearly demonstrate the necessity of upgrading the pre-trained encoder from weak ResNet/CLIP/MiT to strong DINOv2. And the most lightweight DINOv2-S is cheaper to train than ResNet-101. Besides, we use fewer epochs than previous works (60 \emph{vs.} 80 epochs).

\textbf{Cityscapes:} As exhibited in Table~\ref{tab:cityscapes}, across all evaluated splits, our results with the smallest DINOv2-S (24.8M) encoder are superior to all other frameworks based on ResNet-101 (59.5M), MiT-B5 (89.3M), or CLIP (88.0M) encoders. On the 1/16 split, our DINOv2-\emph{\textbf{S}}-based UniMatch V2 improves SemiVL by 2.7\% (77.9\% $\rightarrow$ 80.6\%), and our DINOv2-\emph{\textbf{B}}-based result even outperforms it by 5.7\% (77.9\% $\rightarrow$ 83.6\%). We also report the fully-supervised result in the bracket next to the ``Labeled Only'', where all available labeled images (2975 images) are used for training. Our semi-supervised result with half of the labeled images is almost equal to the fully-supervised upper bound (85.1\% \emph{vs.} 85.2\%). It shows the effectiveness of our semi-supervised algorithm in substantially reducing the annotation cost. It is also worth highlighting that Pascal and Cityscapes are the two easiest benchmarks in semantic segmentation, but the performance improvement is still so tremendous, further revealing the inferiority of previously adopted encoders. We believe evaluating SSS frameworks under the modern DINOv2 encoder will attract more audiences to our field in the future.

\textbf{ADE20K:} Among all existing works, only SemiVL~\cite{semivl} reports their performance on this challenging dataset, with some reproduced results of other methods. As compared in Table~\ref{tab:ade20k}, our DINOv2-\emph{\textbf{B}}-based UniMatch V2 outperforms CLIP-\emph{\textbf{B}}-based SemiVL by nearly 10\% in most settings, \eg, 35.1\% \emph{vs.} 45.0\% (+9.9\%) with 1/32 labeled images, and 39.4\% \emph{vs.} 49.8\% (+10.4\%) with 1/8 labeled images. Moreover, even our DINOv2-\emph{\textbf{S}}-based results are much better than SemiVL on three out of four settings, achieved with 3.5$\times$ fewer model parameters (88.0M \emph{vs.} 24.8M). On such a complex dataset, the advantages of DINOv2-based UniMatch V2 are even more significant. Moreover, we can find there is still a considerable margin between our semi-supervised results (\eg, 49.8\% under 1/8 splits) and the fully-supervised result (54.1\%), indicating there is still much room to further improve our UniMatch V2 results.

\textbf{COCO:} In Table~\ref{tab:coco}, compared with our UniMatch V1~\cite{unimatch} based on the Xception-65~\cite{chollet2017xception} encoder, our lightweight V2 framework improves it remarkably, \eg, 44.4\% $\rightarrow$ 53.2\% on the 1/128 split. Our semi-supervised algorithm also boosts the labeled-only baseline impressively, as large as +10\% (45.8\% $\rightarrow$ 55.8\%) on the 1/256 split, highlighting the effectiveness of our framework and the value of extra unlabeled data. Furthermore, our DINOv2-B-based performance is superior to the best-performed SemiVL~\cite{semivl} on four of five splits, \eg, 56.5\% $\rightarrow$ 63.3\% (+6.8\%) on the 1/32 split. However, we witness the only one inferior result of our V2: on the 1/512 split, SemiVL is better than us by 2.2\% (50.1\% \emph{vs.} 47.9\%), indicating delicate designs may be required under such extremely label-scarce regimes.

\begin{table*}[t]
\setlength\tabcolsep{1.9mm}
    \centering
    \begin{tabular}{lcccccccccccccc}
    \toprule
    
    \multicolumn{5}{c}{\multirow{2}{*}{\textbf{Ablation (DINOv2-B)}}} & \multicolumn{5}{c}{Pascal} & \multicolumn{5}{c}{ADE20K} \\
    
    \cmidrule(lr){6-10}\cmidrule(lr){11-15}
    
    & & & & & 1/16 & 1/8 & 1/4 & 1/2 & Full & 1/64 & 1/32 & 1/16 & 1/8 & 1/4 \\

    Reference & $\mathcal{A}^{img}$ & $\mathcal{A}^{feat}$ & \#Streams & CompDrop & (92) & (183) & (366) & (732) & (1464) & (316) & (631) & (1263) & (2526) & (5052) \\
    
    \midrule

    FixMatch & \checkmark & & 1 & N.A. & 83.8 & 87.4 & 88.2 & 89.9 & 90.2 & 37.8 & 43.2 & 46.4 & 49.5 & 51.0 \\
    
    Figure~\ref{fig:imagefeat1_stream} & \checkmark & \checkmark & 1 & N.A. & 84.7 & 86.6 & 87.7 & 89.6 & 90.2 & 38.1 & 43.6 & 46.0 & 49.8 & 51.3 \\

    Figure~\ref{fig:image2_stream} & \checkmark &  & 2 & N.A. & 86.0 & 86.3 & 88.8 & \textbf{90.1} & 90.3 & 37.3 & 43.7 & 46.3 & 49.6 & 51.8 \\
    
    Figure~\ref{fig:imagefeat2_stream} & \checkmark & \checkmark & 2 & & 82.9 & 86.1 & 88.6 & 89.3 & 90.0 & 37.2 & \textbf{45.0} & 46.4 & \textbf{49.9} & 51.4 \\

    \rowcolor{shadecolor}
    UniMatch V2 & \checkmark & \checkmark & 2 & \checkmark & \textbf{86.3} & \textbf{87.9} & \textbf{88.9} & 90.0 & \textbf{90.8} & \textbf{38.7} & \textbf{45.0} & \textbf{46.7} & 49.8 & \textbf{52.0} \\
    
    \bottomrule
    
    \end{tabular}
    \caption{Ablation study on various choices of learnable streams. $\mathcal{A}^{img}$ denotes image-level augmentations, while $\mathcal{A}^{feat}$ represents feature-level augmentations (\ie, channel-wise Dropout). The ``CompFeat'' is short for complementary Dropout. Our UniMatch V2 stands out as the best design among them.}
    \label{tab:ablation_streams}
\end{table*}

\begin{table*}[t]
\setlength\tabcolsep{1.18mm}
    \centering
    \begin{tabular}{ccccccccccccccccccccc}
    \toprule
    
    \multirow{2}{*}{\textbf{Frozen}} & \multirow{2}{*}{\textbf{Encoder}} & \multicolumn{5}{c}{Pascal} & \multicolumn{4}{c}{Cityscapes} & \multicolumn{5}{c}{ADE20K} & \multicolumn{5}{c}{COCO} \\
    
    \cmidrule(lr){3-7}\cmidrule(lr){8-11}\cmidrule(lr){12-16}\cmidrule(lr){17-21}
    
    & & 1/16 & 1/8 & 1/4 & 1/2 & Full & 1/16 & 1/8 & 1/4 & 1/2 & 1/64 & 1/32 & 1/16 & 1/8 & 1/4 & 1/512 & 1/256 & 1/128 & 1/64 & 1/32 \\
    
    \midrule
    
     & \multirow{3}{*}{DINOv2-S} & 79.0 & 85.5 & 85.9 & 86.7 & 87.8 & 80.6 & 81.9 & 82.4 & 82.6 & 31.5 & 38.1 & 40.7 & 44.4 & 45.8 & 39.3 & 45.4 & 53.2 & 55.0 & 57.0 \\

    \checkmark & & 81.8 & 83.8 & 84.0 & 85.7 & 86.2 & 75.1 & 77.1 & 78.2 & 78.6 & 29.9 & 34.5 & 37.3 & 39.6 & 41.4 & 37.4 & 42.4 & 48.2 & 51.1 & 52.3 \\

    & & \bluetext{\textbf{+2.8}} & \bluetext{\textbf{-1.7}} & \bluetext{\textbf{-1.9}} & \bluetext{\textbf{-1.0}} & \bluetext{\textbf{-1.6}} & \bluetext{\textbf{-5.5}} & \bluetext{\textbf{-4.8}} & \bluetext{\textbf{-4.2}} & \bluetext{\textbf{-4.0}} & \bluetext{\textbf{-1.6}} & \bluetext{\textbf{-3.6}} & \bluetext{\textbf{-3.4}} & \bluetext{\textbf{-4.8}} & \bluetext{\textbf{-4.4}} & \bluetext{\textbf{-1.9}} & \bluetext{\textbf{-3.0}} & \bluetext{\textbf{-5.0}} & \bluetext{\textbf{-3.9}} & \bluetext{\textbf{-4.7}} \\
    
    \midrule

     & \multirow{3}{*}{DINOv2-B} & 86.3 & 87.9 & 88.9 & 90.0 & 90.8 & 83.6 & 84.3 & 84.5 & 85.1 & 38.7 & 45.0 & 46.7 & 49.8 & 52.0 & 47.9 & 55.8 & 58.7 & 60.4 & 63.3 \\
     
    \checkmark & & 84.6 & 87.2 & 87.4 & 88.5 & 90.2 & 79.8 & 81.4 & 81.9 & 82.8 & 35.6 & 40.2 & 42.3 & 45.6 & 47.3 & 43.9 & 50.1 & 54.8 & 57.2 & 58.4 \\

    & & \bluetext{\textbf{-1.7}} & \bluetext{\textbf{-0.7}} & \bluetext{\textbf{-1.5}} & \bluetext{\textbf{-1.5}} & \bluetext{\textbf{-0.6}} & \bluetext{\textbf{-3.8}} & \bluetext{\textbf{-2.9}} & \bluetext{\textbf{-2.6}} & \bluetext{\textbf{-2.3}} & \bluetext{\textbf{-3.1}} & \bluetext{\textbf{-4.8}} & \bluetext{\textbf{-4.4}} & \bluetext{\textbf{-4.2}} & \bluetext{\textbf{-4.7}} & \bluetext{\textbf{-4.0}} & \bluetext{\textbf{-5.7}} & \bluetext{\textbf{-3.9}} & \bluetext{\textbf{-3.2}} & \bluetext{\textbf{-4.9}} \\
    
    \bottomrule
    
    \end{tabular}
    \caption{Ablation study on fine-tuning (by default) or freezing the pre-trained DINOv2 encoder with our UniMatch V2 framework. The frozen practice is 2$\times$ faster than fully fine-tuning in terms of training efficiency.}
    \label{tab:ablation_frozen}
\end{table*}

\subsection{Ablation Studies}

Unless otherwise specified, we conduct all ablation studies under the DINOv2-B encoder, which is more capable than DINOv2-S and thus can provide more room for different designs to show their effectiveness. Distinguished from all existing works that only perform ablation studies on a single dataset with limited splits, we ablate our various designs and choices across extensive splits and datasets. Through these comprehensive results, we hope to shed more light on future works to better improve SSS performance.

\subsubsection{Comparison with FixMatch and UniMatch V1}

Similar to UniMatch V1~\cite{unimatch}, our UniMatch V2 is also built on the most basic FixMatch framework~\cite{fixmatch}. Hence, we primarily present the most important comparison with the FixMatch baseline and our precedent V1 work under the same DINOv2-B encoder in Table~\ref{tab:ablation_baseline}. It is worth noting that, different from their original implementation that uses the online model for pseudo labeling, we re-implement them by maintaining an EMA teacher to produce higher-quality pseudo labels (same as our V2). As exhibited in Table~\ref{tab:ablation_baseline}, despite the strong performance of the reproduced FixMatch, our UniMatch V2 further improves it across all splits of three datasets. For instance, on ADE20K, our UniMatch outperforms FixMatch by 1.8\% (43.2\% $\rightarrow$ 45.0\%) on the 1/32 split, and on COCO, we improve it by 2.1\% (53.7\% $\rightarrow$ 55.8\%) on the 1/256 split. Moreover, compared with UniMatch V1, we not only speed up its training by reducing learnable streams, but also surpass it across almost all the settings (only 1 of 13 settings is 0.1\% inferior), due to our carefully designed unified complementary augmentations. Notably, on the 1/32 split of ADE20K, our V2 is evidently superior to V1 by 2.2\% (42.8\% \emph{vs.} 45.0\%).

\begin{figure}[t]
    \centering
    \hspace{3mm}
    \begin{subfigure}{0.18\linewidth}
        \centering
        \includegraphics[width=\textwidth]{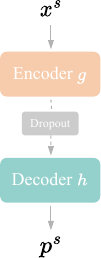}
        \caption{}
        \label{fig:imagefeat1_stream}
    \end{subfigure}
    \hfill
    \begin{subfigure}{0.24\linewidth}
        \centering
        \includegraphics[width=\textwidth]{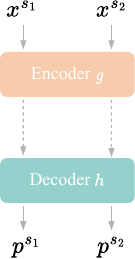}
        \caption{}
        \label{fig:image2_stream}
    \end{subfigure}
    \hfill
    \begin{subfigure}{0.24\linewidth}
        \centering
        \includegraphics[width=\textwidth]{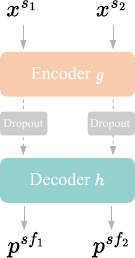}
        \caption{}
        \label{fig:imagefeat2_stream}
    \end{subfigure}
    \hspace{3mm}
    \caption{Other potential designs of the learnable streams. \textbf{(a)} A single unified image-level and feature-level augmentation stream. \textbf{(b)} Two image-level augmentation streams. \textbf{(c)} Two unified image-level and feature-level augmentation streams, using random Dropout rather than our complementary Dropout. See Section~\ref{sec:learnable_streams} and Table~\ref{tab:ablation_streams} for details.}
    \label{fig:learnable_streams}
\end{figure}

\begin{table*}[t]
\setlength\tabcolsep{2.2mm}
    \centering
    \begin{tabular}{lcccccccccccc}
    \toprule

    \multirow{3}{*}{\textbf{Scaling Encoder}} & \multirow{3}{*}{\#Params} & \multirow{3}{*}{Method} & \multicolumn{5}{c}{Pascal} & \multicolumn{5}{c}{ADE20K} \\

    \cmidrule(lr){4-8}\cmidrule(lr){9-13}
    
     & & & 1/16 & 1/8 & 1/4 & 1/2 & Full & 1/64 & 1/32 & 1/16 & 1/8 & 1/4 \\

    & & & (92) & (183) & (366) & (732) & (1464) & (316) & (631) & (1263) & (2526) & (5052) \\
    
    \midrule

    \multirow{3}{*}{DINOv2-Small} & \multirow{3}{*}{24.8M} & Labeled Only & 67.0 & 75.6 & 81.8 & 83.7 & 85.6 & 26.1 & 32.7 & 37.1 & 39.8 & 42.7 \\

    & & \multirow{2}{*}{UniMatch V2} & 79.0 & 85.5 & 85.9 & 86.7 & 87.8 & 31.5 & 38.1 & 40.7 & 44.4 & 45.8 \\

    & &  & \bluetext{\textbf{(+12.0)}} & \bluetext{\textbf{(+9.9)}} & \bluetext{\textbf{(+4.1)}} & \bluetext{\textbf{(+3.0)}} & \bluetext{\textbf{(+2.2)}} & \bluetext{\textbf{(+5.4)}} & \bluetext{\textbf{(+5.4)}} & \bluetext{\textbf{(+3.6)}} & \bluetext{\textbf{(+4.6)}} & \bluetext{\textbf{(+3.1)}} \\
    
    \midrule
    
    \multirow{3}{*}{DINOv2-Base} & \multirow{3}{*}{97.5M} & Labeled Only & 76.9 & 82.1 & 85.3 & 87.2 & 88.3 & 32.1 & 39.3 & 42.8 & 46.4 & 49.0 \\

    & & \multirow{2}{*}{UniMatch V2} & 86.3 & 87.9 & 88.9 & 90.0 & 90.8 & 38.7 & 45.0 & 46.7 & 49.8 & 52.0 \\

    & &  & \bluetext{\textbf{(+9.4)}} & \bluetext{\textbf{(+5.8)}} & \bluetext{\textbf{(+3.6)}} & \bluetext{\textbf{(+2.8)}} & \bluetext{\textbf{(+2.5)}} & \bluetext{\textbf{(+6.6)}} & \bluetext{\textbf{(+5.7)}} & \bluetext{\textbf{(+3.9)}} & \bluetext{\textbf{(+3.4)}} & \bluetext{\textbf{(+3.0)}} \\
    
    \midrule

    \multirow{3}{*}{DINOv2-Large} & \multirow{3}{*}{335.6M} & Labeled Only & 76.9 & 82.4 & 86.8 & 89.4 & 90.7 & 33.2 & 39.9 & 45.0 & 49.2 & 51.8 \\

    & & \multirow{2}{*}{UniMatch V2} & 86.4 & 87.6 & 90.3 & 91.1 & 91.2 & 40.0 & 45.0 & 49.3 & 52.1 & 54.2 \\
    
    & & & \bluetext{\textbf{(+9.5)}} & \bluetext{\textbf{(+5.2)}} & \bluetext{\textbf{(+3.5)}} & \bluetext{\textbf{(+1.7)}} & \bluetext{\textbf{(+0.5)}} & \bluetext{\textbf{(+6.8)}} & \bluetext{\textbf{(+5.1)}} & \bluetext{\textbf{(+4.3)}} & \bluetext{\textbf{(+2.9)}} & \bluetext{\textbf{(+2.4)}} \\
    
    \bottomrule
    
    \end{tabular}
    \caption{Ablation study on scaling up the encoder capacity. Updating the encoder from DINOv2-B to a larger DINOv2-L can further improve our UniMatch V2 results. Across three encoder scales, we all demonstrate the effectiveness of our proposed framework in leveraging unlabeled images.}
    \label{tab:scaling_encoder}
\end{table*}

\begin{table}[t]
\setlength\tabcolsep{2.4mm}
    \centering
    \begin{tabular}{lccccc}
    \toprule
    
    \multirow{3}{*}{\textbf{Encoders}} & \multirow{3}{*}{\#Params}  & \multicolumn{3}{c}{ADE20K} & \multicolumn{1}{c}{Pascal} \\
    
    \cmidrule(lr){3-5}\cmidrule(lr){6-6}

    & & 1/64  & 1/16 & 1/4 & 1/8 \\

    & & (316) & (1263) & (5052) & (183) \\
    
    \midrule
    
    ResNet-152~\cite{resnet} & 78.6M & 23.2 & 32.2 & 37.2 & 76.2 \\

    \rowcolor{shadecolor}
    DINOv2-S~\cite{dinov2} & 24.8M & \textbf{31.5} & \textbf{40.7} & \textbf{45.8} & \textbf{85.5} \\
    
    \midrule

    SAM-B~\cite{sam} & 99.8M & 21.4 & 31.5 & 36.1 & 63.2 \\
    
    MiT-B5~\cite{segformer} & 99.2M & 27.9 & 37.4 & 42.8 & 81.4 \\
    
    BEiT-B~\cite{beit} & 97.2M & 29.4 & 39.6 & 44.2 & 84.5 \\
    
    \rowcolor{shadecolor}
    DINOv2-B~\cite{dinov2} & 97.5M & \textbf{38.7} & \textbf{46.7} & \textbf{52.0} & \textbf{87.9} \\
    
    \midrule

    SAM-L~\cite{sam} & 338.6M & 30.6 & 39.4 & 44.0 & 77.4 \\
    
    BEiT-L~\cite{beit} & 335.9M & 34.6 & 43.7 & 50.1 & 85.1 \\

    \rowcolor{shadecolor}
    DINOv2-L~\cite{dinov2} & 335.6M & \textbf{40.0} & \textbf{49.3} & \textbf{54.2} & \textbf{87.6} \\
    
    \bottomrule
    
    \end{tabular}
    \caption{Ablation study on the capability of various pre-trained encoders with our UniMatch V2 framework. In all cases, we cascade a DPT decoder~\cite{dpt} upon the encoder. We count the total parameters of the encoder and the decoder.}
    \label{tab:ablation_encoder}
\end{table}

\subsubsection{\label{sec:learnable_streams}Other Variants of the Learnable Stream Design}

We design two learnable streams with image-level augmentations and complementary Dropout to effectively harness unlabeled images. To convincingly demonstrate the superiority and necessity of such a design, we further explore three alternative designs on the learnable streams, as shown in Figure~\ref{fig:learnable_streams}. Concretely, we attempt to 1) only use a \emph{single} unified augmentation stream with image augmentations and a random Dropout (Figure~\ref{fig:imagefeat1_stream}), 2) use dual-stream image augmentations, \emph{without feature augmentations} (Figure~\ref{fig:image2_stream}), and 3) use dual-stream unified augmentations, but adopting two \emph{independent} Dropouts, instead of our carefully designed complementary Dropout (Figure~\ref{fig:imagefeat2_stream}). As comprehensively validated in Table~\ref{tab:ablation_streams}, the design of our UniMatch V2 is more effective than all other counterparts. Notably, on Pascal with 183 labeled images, we outperform the other three designs by 1.3\%, 1.6\%, and 1.8\%, respectively.

\subsubsection{Fine-tuning vs. Freezing the DINOv2 Encoder}

Unless otherwise specified in this paper, we fully fine-tune the entire model (\ie, encoder + decoder). However, limited by the GPU memory, some academic groups cannot afford such a training strategy. In view of this, we further attempt to freeze the pre-trained DINOv2 encoder and solely train the randomly initialized DPT decoder, which only accounts for 10\% of the total parameters. This greatly reduces the training cost, \eg, GPU memory consumption (DINOv2-B on Pascal) is reduced from 23G to 10G, and training time is reduced by 59\%. The performance of fine-tuning or freezing the encoder is listed in Table~\ref{tab:ablation_frozen}. It is within expectation that fine-tuning achieves better results than freezing the encoder. Among all the 38 settings (spanning various datasets, splits, encoders), only on the Pascal 1/16 split with DINOv2-S, fine-tuning the encoder is inferior to freezing it, with a gap of 2.8\% (79.0\% \emph{vs.} 81.8\%). For the remaining 37 settings, fine-tuning clearly exceeds freezing by 0.6\% - 5.5\%.

Carefully analyzing the gap between fine-tuning and freezing, we can find their gap is larger on challenging datasets, such as ADE20K and COCO, with an average gap of around 3.5\%, while on the easiest Pascal, the average gap is only 1.4\%. Although Cityscapes contains even fewer classes than Pascal, the gap on it is much larger, indicating that there may exist a larger distribution shift between the DINOv2 pre-training set and Cityscapes. Lastly, we want to highlight that even the frozen smallest DINOv2-S can remarkably outperform previous fine-tuned larger ResNet-101 on Pascal, ADE20K, and COCO, also with a reduction in training cost (GPU memory $\downarrow$ 47\%). Motivated by the strong results achieved by a frozen encoder, we believe future works can come up with better strategies (\eg, LoRA~\cite{lora}) to utilize the powerful encoder.

\subsubsection{Scaling Up the Capacity of the DINOv2 Encoder}

Till now, we have reported many results under the DINOv2-S and DINOv2-B encoders. In practice, they are sufficient to obtain promising results. As far as we know, the largest models adopted by previous SSS works~\cite{semivl, allspark} contain nearly 100M parameters, similar to our DINOv2-B-based DPT. Nevertheless, we are still curious whether the SSS results can be further improved by simply scaling up the capacity of the segmentation model. Therefore, we evaluate UniMatch V2 with a DINOv2-\emph{\textbf{L}}-based DPT model, containing 335.6M parameters (3.5$\times$ larger than DINOv2-B). In Table~\ref{tab:scaling_encoder}, we reveal that our results can be further enhanced by updating the encoder from DINOv2-B to DINOv2-L. For example, on the 1/16 split of ADE20K, our result is improved from 46.7\% $\rightarrow$ 49.3\% (+2.6\%). Additionally, even with the strongest DINOv2-L encoder, our UniMatch V2 still consistently improves the labeled-only results, showcasing its advantages in utilizing unlabeled images.

\subsubsection{Comparison with Other Pre-trained Encoders}

We here compare our adopted DINOv2 with other capable pre-trained vision encoders, including ResNet-152~\cite{resnet}, MiT-B5~\cite{segformer}, BEiT-B~\cite{beit}, BEiT-L~\cite{beit}, SAM-B~\cite{sam}, and SAM-L~\cite{sam}. It is worth noting that we carefully re-find an optimal learning rate for these encoders, instead of directly using our DINOv2's learning rate. We apply the same UniMatch V2 framework for these encoders, except BEiT. When adopting BEiT, we use the student model to produce pseudo labels, because we find its EMA teacher is very poor, probably due to the relative positional encoding.

As shown in Table~\ref{tab:ablation_encoder}, we divide all encoders into three groups based on their number of parameters. There are four key observations. (1) Within each group, our adopted DINOv2 encoder significantly outperforms other alternatives. Our DINOv2-S is better than ResNet-152 by 8\% on average, also requiring 3$\times$ fewer parameters. Compared with the strong BEiT-L~\cite{beit} encoder on ADE20K, our DINOv2-L results are 5\% higher on average of three splits. (2) Our results under smaller DINOv2 are even better than other larger models. DINOv2-S consistently surpasses BEiT-B, and DINOv2-B is superior to BEiT-L. (3) Although SAM~\cite{sam} is specifically designed for image segmentation, its encoder performs poorly in our SSS task, mainly due to its weak semantic capability acquired by the class-agnostic segmentation pre-training task. (4) Some prior works fail to fully unlock the capability of advanced encoders. E.g., the results of AllSpark~\cite{allspark} (Table~\ref{tab:pascal_origin}) are much inferior to our UniMatch V2 under the same MiT-B5 encoder (Table~\ref{tab:ablation_encoder}), \eg, 78.4\% \emph{vs.} 81.4\% on the Pascal 183 split.

\begin{figure*}[t]
    \centering
    \begin{subfigure}{0.23\textwidth}
        \centering
        \includegraphics[width=\textwidth]{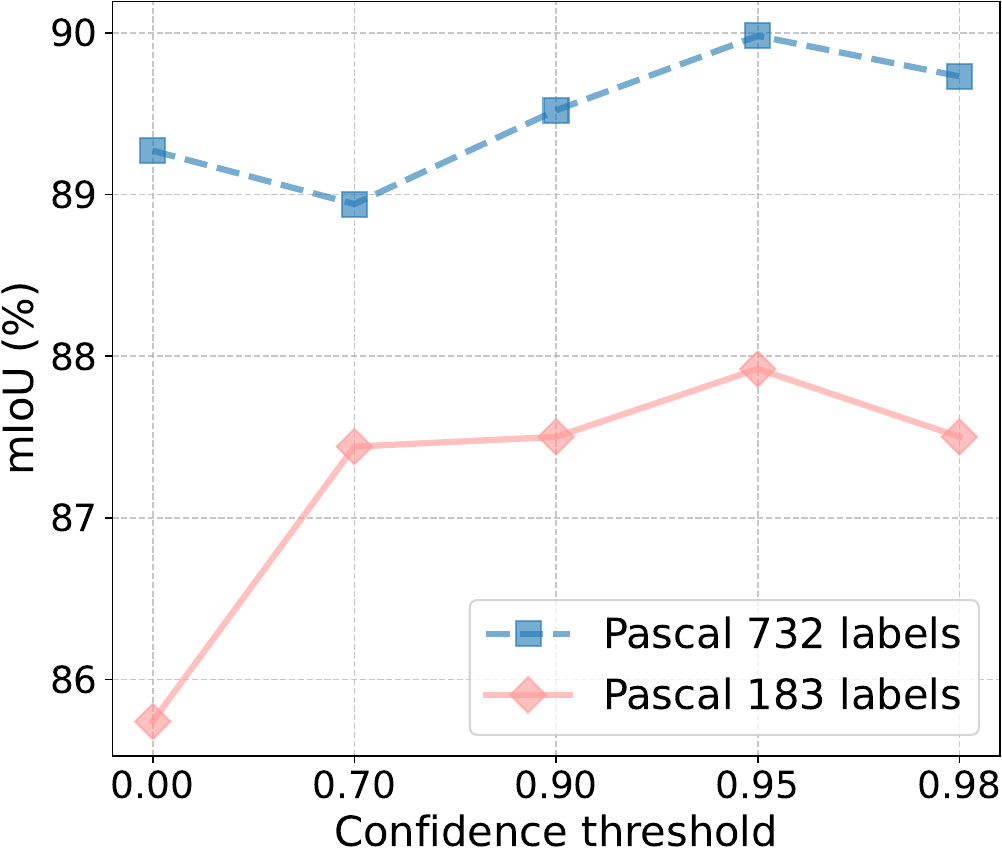}
        \caption{Pascal}
        \label{fig:confidence_pascal}
    \end{subfigure}
    \hfill
    \begin{subfigure}{0.23\textwidth}
        \centering
        \includegraphics[width=\textwidth]{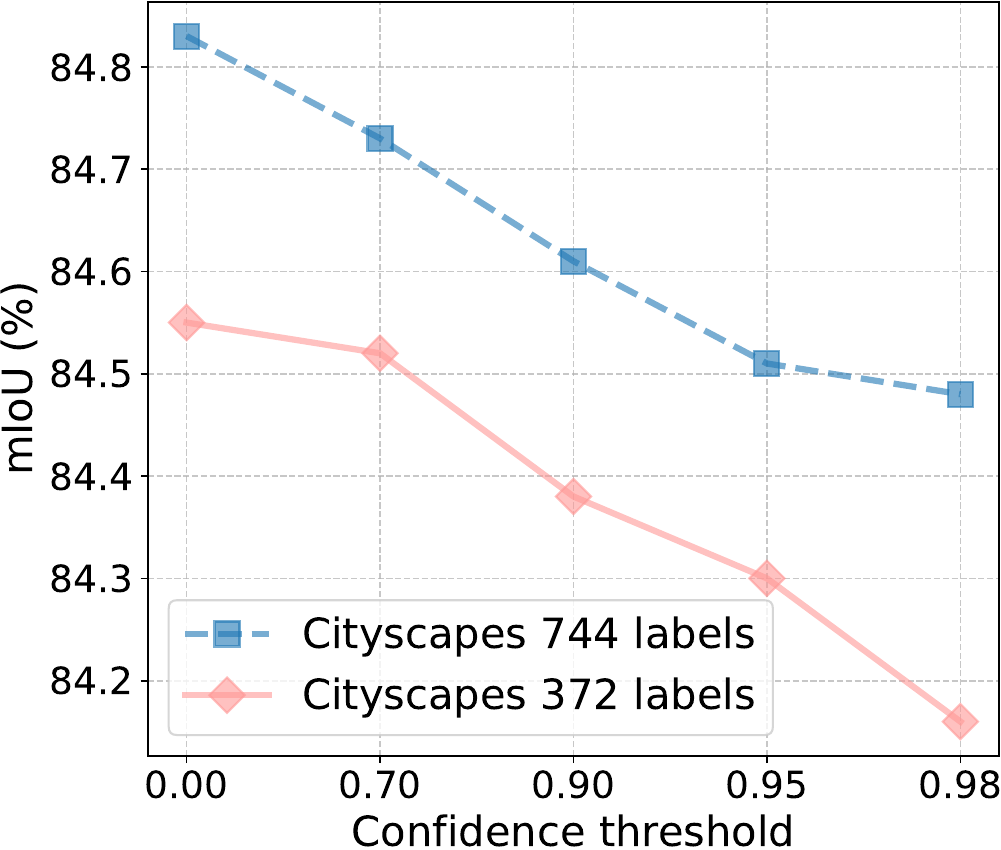}
        \caption{Cityscapes}
        \label{fig:confidence_cityscapes}
    \end{subfigure}
    \hfill
    \begin{subfigure}{0.23\textwidth}
        \centering
        \includegraphics[width=\textwidth]{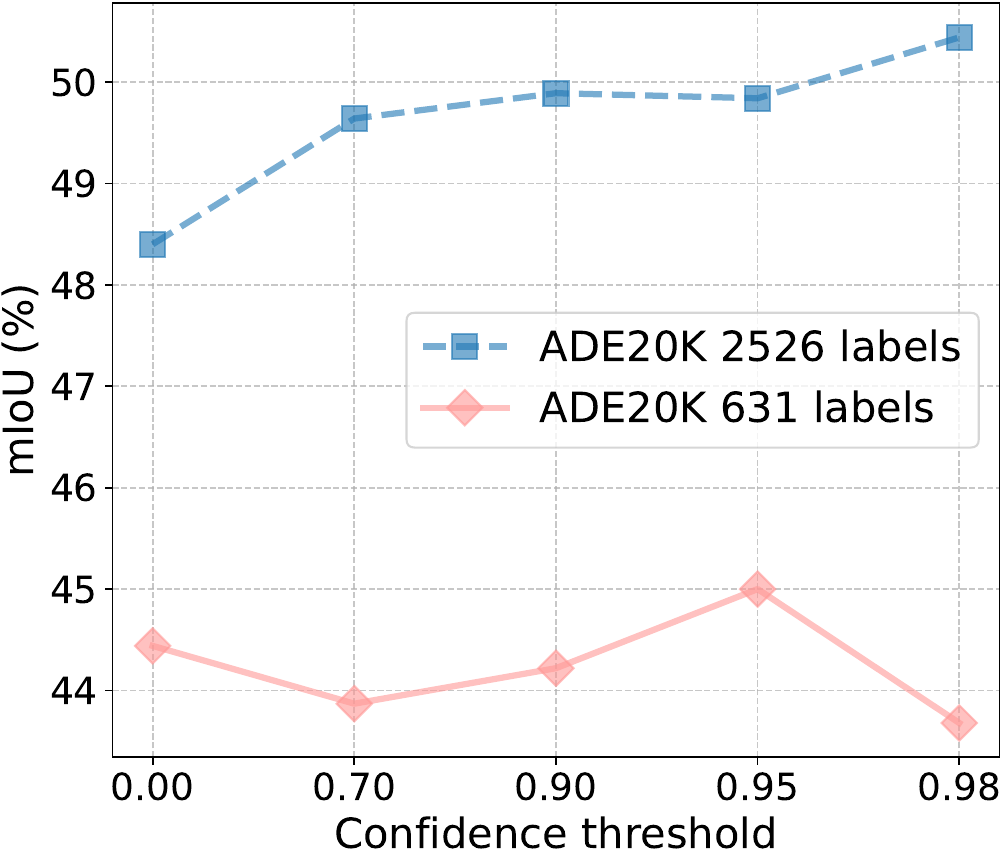}
        \caption{ADE20K}
        \label{fig:confidence_ade20k}
    \end{subfigure}
    \hfill
    \begin{subfigure}{0.23\textwidth}
        \centering
        \includegraphics[width=\textwidth]{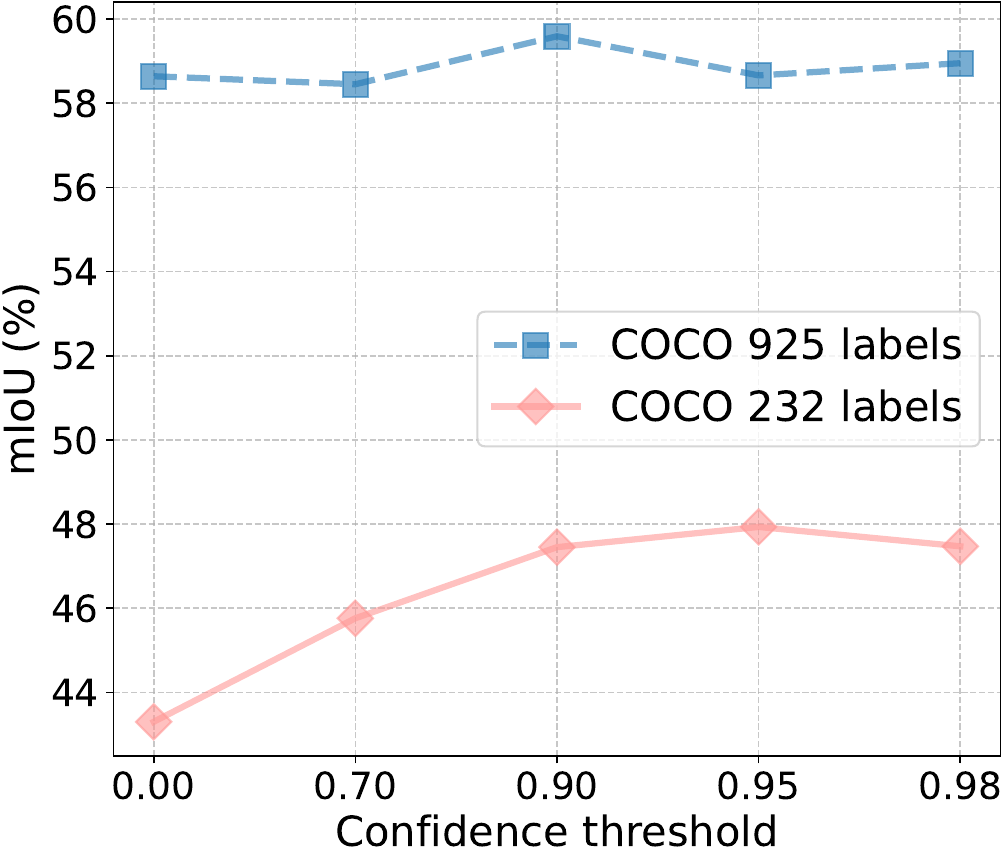}
        \caption{COCO}
        \label{fig:confidence_coco}
    \end{subfigure}
    
    \caption{Ablation study on the confidence threshold $\tau$ (0.95 by default) used to select high-quality pseudo labels.}
    \label{fig:confidence}
\end{figure*}

\begin{figure*}[t]
    \centering
    \begin{subfigure}{0.23\textwidth}
        \centering
        \includegraphics[width=\textwidth]{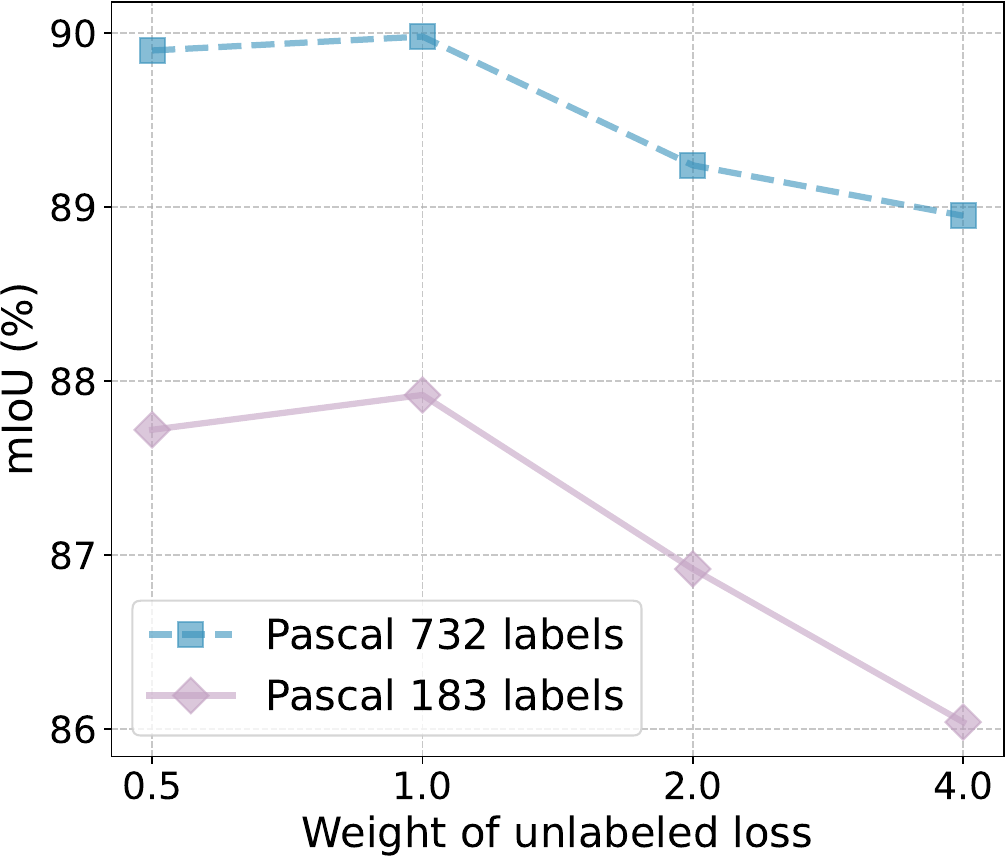}
        \caption{Pascal}
        \label{fig:unsup_lossw_pascal}
    \end{subfigure}
    \hfill
    \begin{subfigure}{0.23\textwidth}
        \centering
        \includegraphics[width=\textwidth]{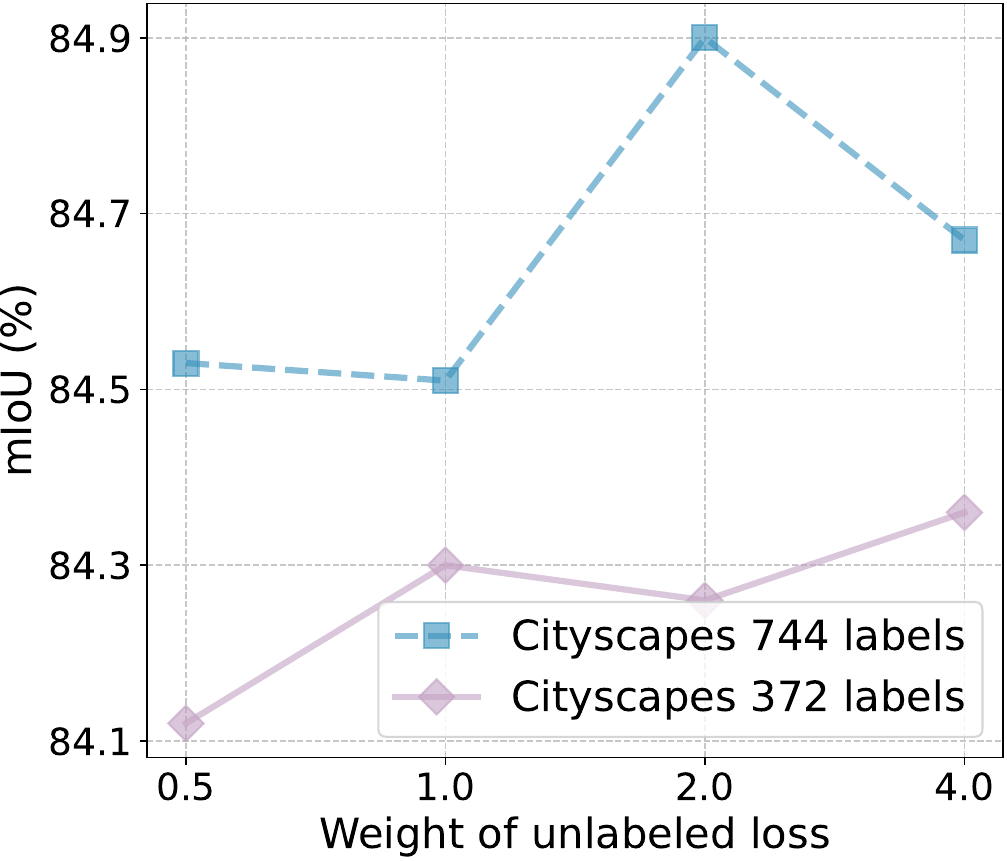}
        \caption{Cityscapes}
        \label{fig:unsup_lossw_cityscapes}
    \end{subfigure}
    \hfill
    \begin{subfigure}{0.23\textwidth}
        \centering
        \includegraphics[width=\textwidth]{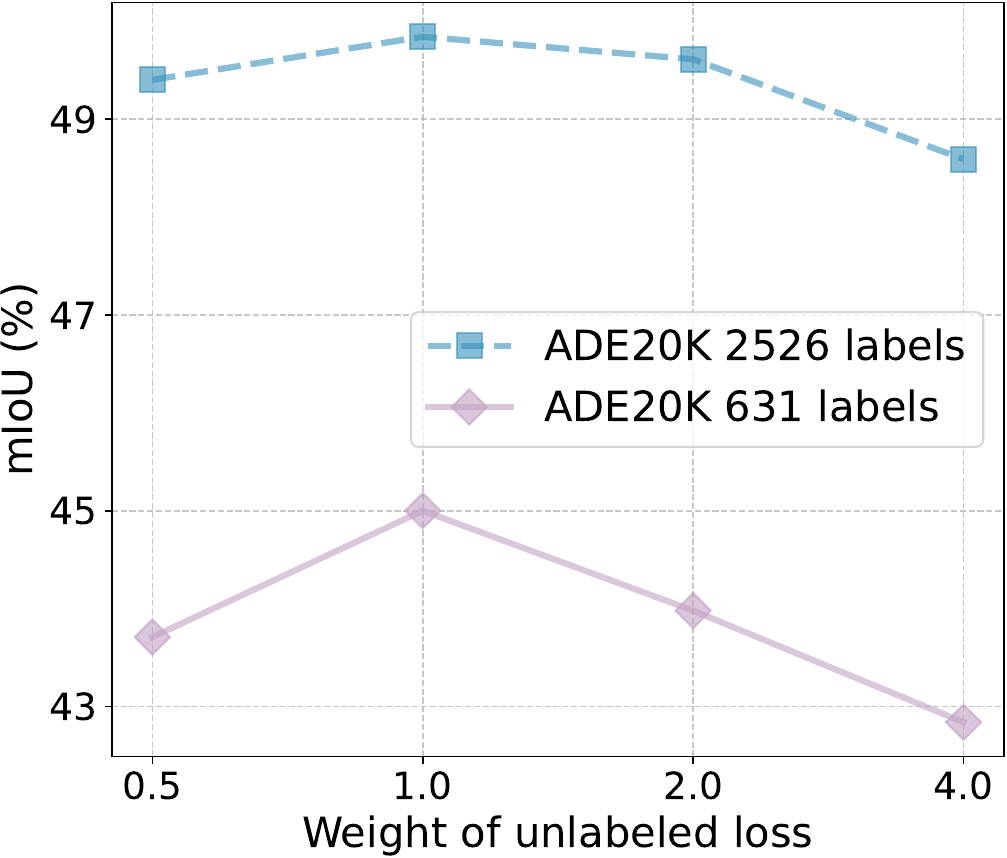}
        \caption{ADE20K}
        \label{fig:unsup_lossw_ade20k}
    \end{subfigure}
    \hfill
    \begin{subfigure}{0.23\textwidth}
        \centering
        \includegraphics[width=\textwidth]{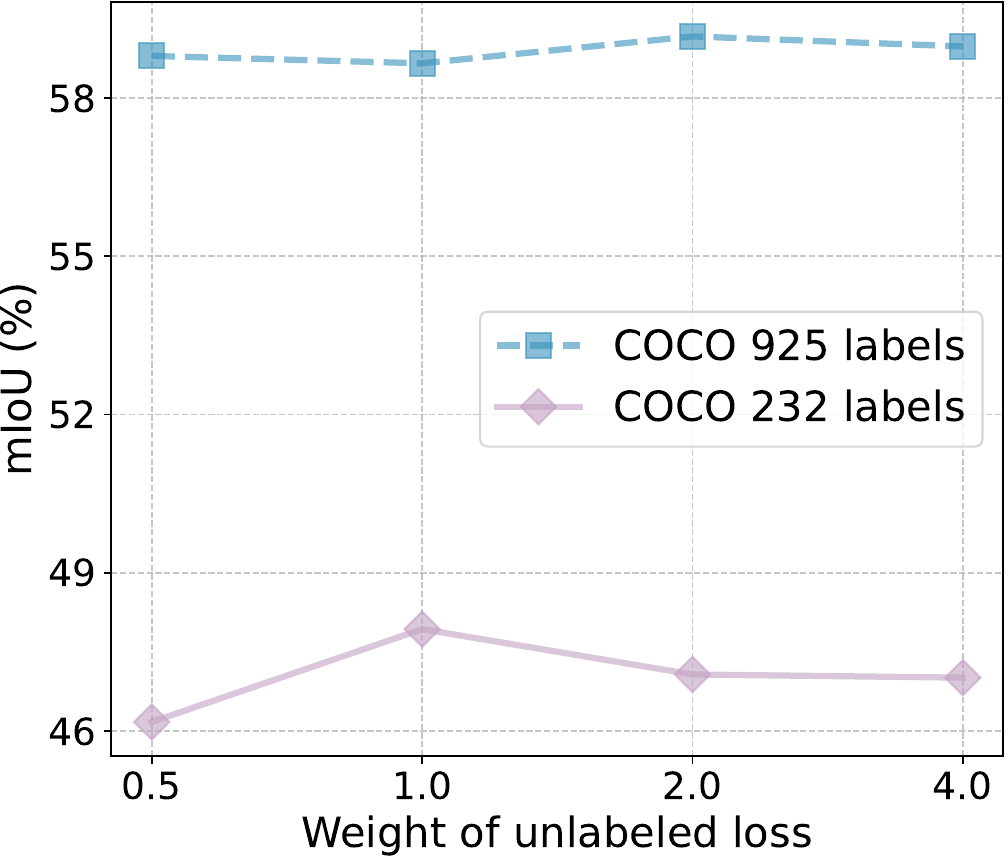}
        \caption{COCO}
        \label{fig:unsup_lossw_coco}
    \end{subfigure}
    
    \caption{Ablation study on the weight $\lambda$ of unlabeled loss $\mathcal{L}^u$. The default unlabeled loss weight 1.0 yields the best results.}
    \label{fig:unlabeled_loss_weight}
\end{figure*}

\subsubsection{Values of the Confidence Threshold}

As aforementioned, we follow FixMatch~\cite{fixmatch} to use a confidence threshold $\tau$ to discard low-confident noisy pseudo labels. By default, we set it as 0.95 for all settings. Here, we ablate its different values across all the four evaluated datasets with two labeled splits each. As exhibited in Figure~\ref{fig:confidence}, we try five values of $\tau$: 0 (not filtering), 0.7, 0.90, 0.95 (default), and 0.98. It can be concluded that in many cases (4 out of 8 settings), 0.95 is already the optimal value for $\tau$ and yields the most consistent performance. A clear exception is the Cityscapes dataset (Figure~\ref{fig:confidence_cityscapes}), where there is a negative correlation between the value of $\tau$ and model performance: the larger $\tau$ is, the lower performance is. The optimal $\tau$ for Cityscapes is 0, which means all pseudo labels are used for training without filtering. This is indeed the same value set by UniMatch V1 for Cityscapes. However, in this V2 work, we avoid specifically fine-tuning this value for each dataset and maintain the same $\tau$ for all datasets.

\subsubsection{Values of the Weight $\lambda$ of Unlabeled Loss }

By default, we set the weight $\lambda$ of unlabeled loss $\mathcal{L}^u$ as 1.0, the same as that of labeled loss. Here we ablate other values (\ie, 0.5, 2.0, 4.0) of the unlabeled loss weight. As shown in Figure~\ref{fig:unlabeled_loss_weight}, the default weight 1.0 is almost the optimal value, better than all other weights in 6 out of 8 settings (spanning all the four datasets, each with two labeled data splits). In most cases, when the loss weight has exceeded 1.0, the larger the value is, the worse performance will be.

\begin{table}[t]
\setlength\tabcolsep{1.5mm}
    \centering
    \begin{tabular}{ccccccccc}
    \toprule
    
    \multirow{3}{*}{\textbf{EMA Teacher}} & \multicolumn{2}{c}{Pascal} & \multicolumn{2}{c}{ADE20K} & \multicolumn{3}{c}{COCO} \\

    \cmidrule(lr){2-3}\cmidrule(lr){4-5}\cmidrule(lr){6-8}
    
    & 1/16 & 1/8 & 1/64 & 1/32 & 1/512 & 1/256 & 1/128 \\

    & (92) & (183) & (316) & (631) & (232) & (463) & (925) \\
    
    \midrule

    \checkmark & \textbf{86.3} & \textbf{87.9} & \textbf{38.7} & \textbf{45.0} & \textbf{47.9} & \textbf{55.8} & \textbf{58.7} \\
    
    & 83.5 & 86.5 & 37.7 & 42.8 & 44.9 & 54.1 & 57.6 \\
    
    \bottomrule
    
    \end{tabular}
    \caption{Ablation study on whether to use an EMA teacher or directly use the online student to produce pseudo label.}
    \label{tab:ablation_ema}
\end{table}

\begin{figure}[t]
    \centering
    \begin{subfigure}{0.322\linewidth}
        \centering
        \includegraphics[width=\textwidth]{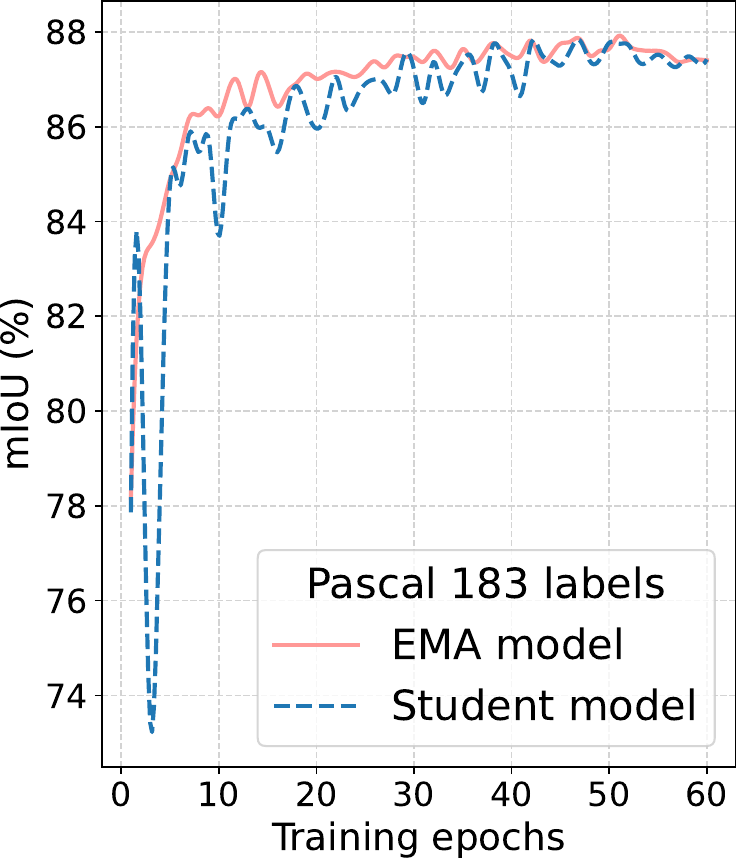}
        \caption{Pascal}
        \label{fig:ema_results_pascal}
    \end{subfigure}
    \hfill
    \begin{subfigure}{0.32\linewidth}
        \centering
        \includegraphics[width=\textwidth]{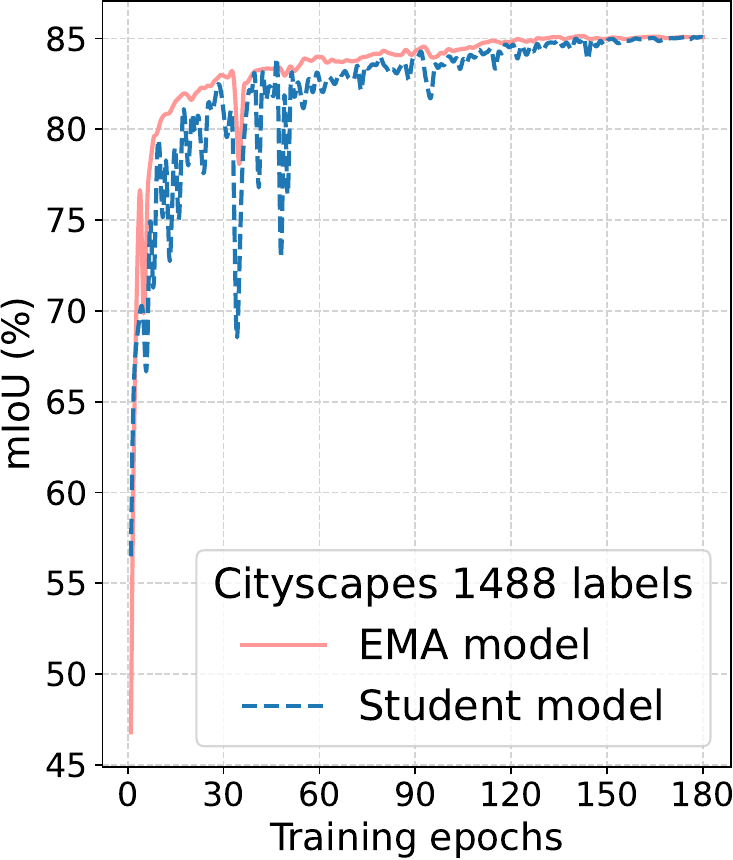}
        \caption{Cityscapes}
        \label{fig:ema_results_cityscapes}
    \end{subfigure}
    \hfill
    \begin{subfigure}{0.322\linewidth}
        \centering
        \includegraphics[width=\textwidth]{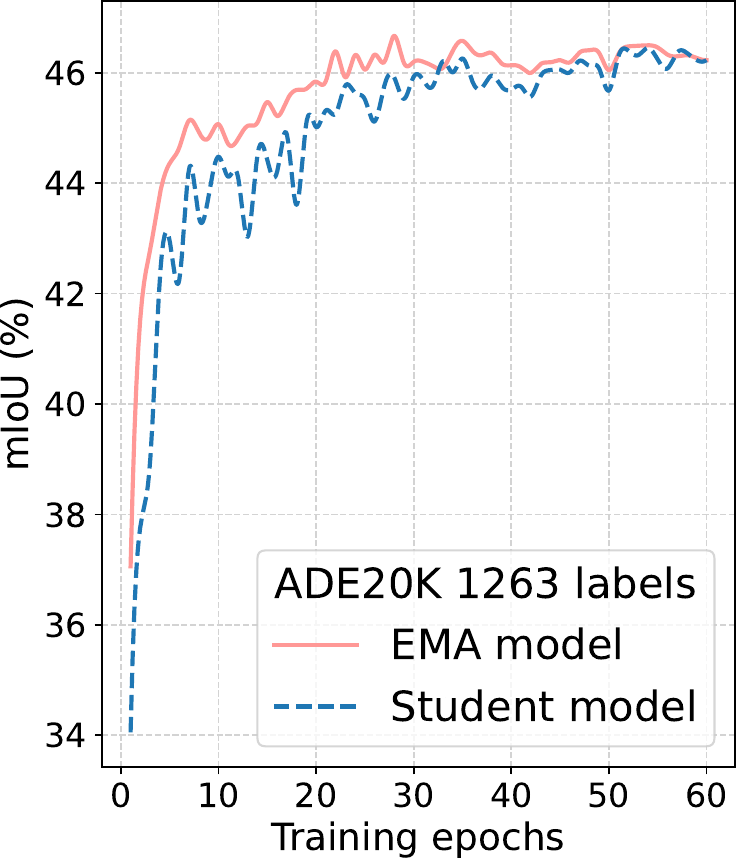}
        \caption{ADE20K}
        \label{fig:ema_results_ade20k}
    \end{subfigure}
    
    \caption{Comparison between the performance of the online trained student model and the EMA teacher model.}
    \label{fig:ema_results}
\end{figure}

\subsubsection{Effectiveness of the EMA Teacher}

Different from FixMatch and our previous UniMatch V1, we use an EMA teacher to produce pseudo labels for the online student to learn in V2. This modification is motivated by observations that the EMA model is mostly 1\% - 2\% better than the online model at early training stages and its performance is more stable, as visualized in Figure~\ref{fig:ema_results}. In Table~\ref{tab:ablation_ema}, we further compare the final performance of using the EMA teacher \emph{vs.} using the online student to produce pseudo labels. In the former case (our default practice), we report the EMA teacher test results. In the latter case, we report better test results of the online model and its EMA version. For all evaluated settings of three datasets, using the EMA model yields much better results, further proving its higher labeling quality.

\subsubsection{Dual-Stream Learning vs. Doubled Batch Size}

With the design of dual complementary streams, our UniMatch V2 is significantly superior to FixMatch. However, this also comes with nearly doubled training cost. Therefore, we need examine whether FixMatch can be on par with ours if training it with doubled batch size for doubled epochs, \ie, whether our better results can be trivially achieved by simply increasing the training budget. As shown in Table~\ref{tab:ablation_doubled_bs}, although we manually increase the training cost of FixMatch by two times, it is still evidently inferior to our UniMatch V2 on the three datasets, showcasing the effectiveness of our designed complementary views.

\subsubsection{Inserted Position of the Complementary Dropout}

Following UniMatch V1, by default, we insert our proposed complementary Dropout at the intersection of the encoder (\eg, DINOv2) and the decoder (\eg, DPT head). We also tried to move this mechanism to the intersection of the decoder and the final convolutional classifier. Although it is inserted at the back of the model, the gradient will be backpropagated throughout the entire model, enhancing the overall robustness. As compared in Table~\ref{tab:ablation_dropout_position}, between the two attempted positions, the encoder-decoder position is generally better than the decoder-classifier position, same as the observation in UniMatch V1. It demonstrates that it is more beneficial to inject feature-level augmentations in the middle of the model.

\begin{table}[t]
\setlength\tabcolsep{1.9mm}
    \centering
    \begin{tabular}{lccccc}
    \toprule

    \multirow{2}{*}{Method} & \multicolumn{2}{c}{Pascal} & \multicolumn{2}{c}{ADE20K} & COCO \\

    \cmidrule(lr){2-3}\cmidrule(lr){4-5}\cmidrule(lr){6-6}
    
     & 1/16 & 1/8 & 1/64 & 1/32 & 1/512\\
     
    \midrule

    FixMatch + 2$\times$ BS \& Epoch & 84.4 & 86.7 & 36.8 & 44.4 & 47.5 \\

    Our UniMatch V2 & \textbf{86.3} & \textbf{87.9} & \textbf{38.7} & \textbf{45.0} & \textbf{47.9} \\
    
    \bottomrule
    
    \end{tabular}
    \caption{Our UniMatch V2, designed with complementary dual streams, is superior to the FixMatch naively augmented by doubled batch size and doubled epochs.}
    \label{tab:ablation_doubled_bs}
\end{table}

\begin{table}[t]
\setlength\tabcolsep{1.4mm}
    \centering
    \begin{tabular}{lccccccc}
    \toprule

    \multirow{2}{*}{\textbf{Dropout Position}} & \multicolumn{4}{c}{Pascal} & \multicolumn{2}{c}{ADE20K} & \multicolumn{1}{c}{COCO} \\

    \cmidrule(lr){2-5}\cmidrule(lr){6-7}\cmidrule(lr){8-8}
    
    & 1/16 & 1/8 & 1/4 & 1/2 & 1/32 & 1/16 & 1/256 \\

    \midrule

    Encoder - Decoder & \textbf{86.3} & \textbf{87.9} & \textbf{88.9} & \textbf{90.0} & \textbf{45.0} & 46.7 & \textbf{55.8} \\

    Decoder - Classifier & 83.5 & 87.1 & \textbf{88.9} & 89.7 & 43.5 & \textbf{46.9} & 54.6 \\
    
    \bottomrule
    
    \end{tabular}
    \caption{Ablation study on different inserted positions of our complementary Dropout.}
    \label{tab:ablation_dropout_position}
\end{table}

\begin{figure}[t]
    \centering
    \includegraphics[width=0.85\linewidth]{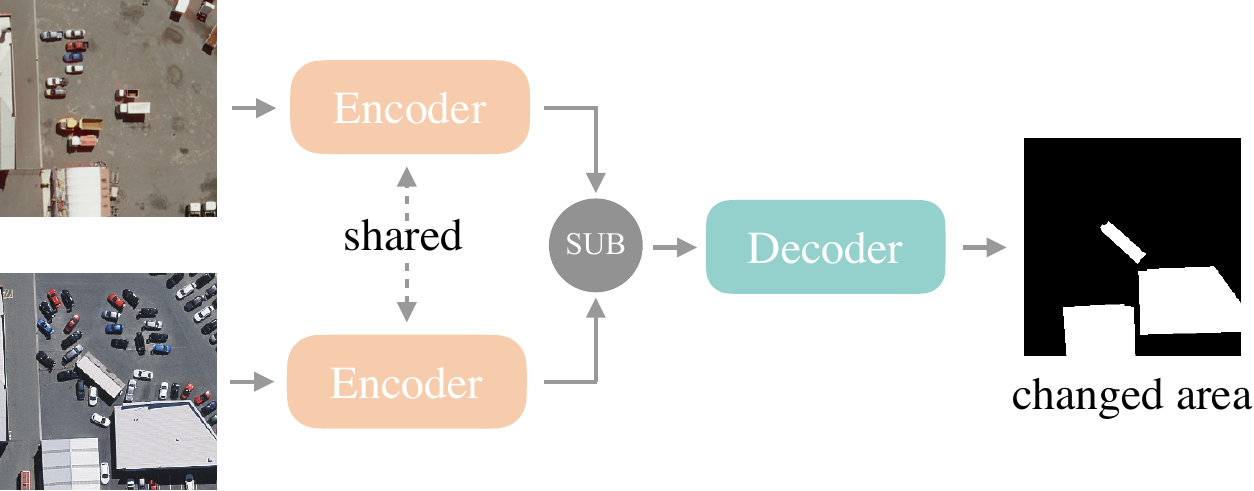}
    \caption{A widely adopted pipeline for remote sensing change detection. Features extracted from bi-temporal images are subtracted and then sent into the decoder for binary segmentation.}
    \label{fig:cd_pipeline}
\end{figure}

\begin{table}[t]
\setlength\tabcolsep{2mm}
    \centering
    \begin{tabular}{llc}
    \toprule
    
    Labeled Data (\# Img) & + Unlabeled Data (\# Img) & Improvement \\
    
    \midrule

    COCO (118K) & COCO Extra (123K) & 66.4 $\rightarrow$ 67.1 \\
    
    ADE20K (20K) & COCO Labeled (118K) & 54.1 $\rightarrow$ 54.9 \\

    ADE20K (20K) & COCO All (118K + 123K) & \textbf{54.1 $\rightarrow$ 55.7} \\
    
    Cityscapes (3K) & Cityscapes Extra (20K) & 85.2 $\rightarrow$ 85.5 \\
    
    \bottomrule
    
    \end{tabular}
    \caption{Making efforts towards a \emph{real-world} and \emph{large-scale} semi-supervised semantic segmentation setting, where many images (\eg, 10K) have been labeled over time, and meantime larger-scale unlabeled images (\eg, 100K) are available.}
    \label{tab:large_scale_sss}
\end{table}

\begin{table*}[t]
\setlength\tabcolsep{2.23mm}
    \centering
    \begin{tabular}{lccccccccccccc}
    \toprule

    \multirow{3}{*}{\textbf{Change Detection}} & \multirow{3}{*}{\textbf{Model}} & \multicolumn{8}{c}{LEVIR-CD} & \multicolumn{4}{c}{WHU-CD} \\

    \cmidrule(lr){3-10}\cmidrule(lr){11-14}
    
    & & \multicolumn{2}{c}{5\% (356)} & \multicolumn{2}{c}{10\% (712)} & \multicolumn{2}{c}{20\% (1424)} & \multicolumn{2}{c}{40\% (2848)} & \multicolumn{2}{c}{20\% (1189)} & \multicolumn{2}{c}{40\% (2378)} \\

    \cmidrule(lr){3-4}\cmidrule(lr){5-6}\cmidrule(lr){7-8}\cmidrule(lr){9-10}\cmidrule(lr){11-12}\cmidrule(lr){13-14}
    
    & & IoU$^c$ & OA & IoU$^c$ & OA & IoU$^c$ & OA & IoU$^c$ & OA & IoU$^c$ & OA & IoU$^c$ & OA \\
    
    \midrule
    
    AdvNet~\cite{vu2019advent} & FCN-RN-50 & 66.1 & 98.08 & 72.3 & 98.45 & 74.6 & 98.58 & 75.0 & 98.60 & 73.8 & 98.80 & 76.6 & 98.94 \\
    
    S4GAN~\cite{mittal2019semi} & FCN-RN-50 & 64.0 & 97.89 & 67.0 & 98.11 & 73.4 & 98.51 & 75.4 & 98.62 & 70.8 & 98.60 & 76.4 & 98.96 \\

    SemiCDNet~\cite{peng2020semicdnet} & UNet++ & 67.6 & 98.17 & 71.5 & 98.42 & 74.3 & 98.58 & 75.5 & 98.63 & 66.7 & 98.28 & 75.9 & 98.93 \\

    SemiCD~\cite{bandara2022revisiting} & FCN-RN-50 & 72.5 & 98.47 & 75.5 & 98.63 & 76.2 & 98.68 & 77.2 & 98.72 & 74.8 & 98.84 & 77.2 & 98.96 \\
    
    \textbf{UniMatch V1}~\cite{unimatch} & PSPNet-RN-50 & 75.6 & 98.62 & 79.0 & 98.83 & 79.0 & 98.84 & 78.2 & 98.79 & 82.9 & 99.26 & 84.4 & 99.32 \\
    
    \textbf{UniMatch V1}~\cite{unimatch} & DeepLab-RN-50 & 80.7 & 98.95 & 82.0 & 99.02 & 81.7 & 99.02 & 82.1 & 99.03 & 81.7 & 99.18 & 85.1 & 99.35 \\

    SemiCD-VL~\cite{li2024diffmatch} & RN-50 + VLM & 81.9 & 99.02 & 82.6 & 99.06 & 82.7 & 99.05 & 83.0 & 99.07 & 84.8 & 99.36 & 85.7 & 99.39 \\
    
    \midrule

    \rowcolor{shadecolor}
    Labeled Only & & 78.1 & 98.77 & 80.4 & 98.91 & 81.1 & 98.95 & 82.4 & 99.03 & 79.5 & 99.09 & 82.6 & 99.25 \\

    \rowcolor{shadecolor}
    \textbf{UniMatch V2} & \multirow{-2}{*}{DPT-DINOv2-S} & 82.1 & 99.01 & 82.7 & 99.05 & 83.2 & 99.08 & 83.5 & 99.10 & 86.1 & 99.42 & 87.6 & 99.48 \\

    \midrule

    \rowcolor{shadecolor}
    Labeled Only & & 79.5 & 98.86 & 81.3 & 98.96 & 82.4 & 99.02 & 83.6 & 99.10 & 83.3 & 99.27 & 87.0 & 99.45 \\

    \rowcolor{shadecolor}
    \textbf{UniMatch V2} & \multirow{-2}{*}{DPT-DINOv2-B} & \textbf{83.3} & \textbf{99.08} & \textbf{83.8} & \textbf{99.11} & \textbf{84.3} & \textbf{99.14} & \textbf{84.3} & \textbf{99.14} & \textbf{87.9} & \textbf{99.50} & \textbf{88.6} & \textbf{99.52} \\
    
    \bottomrule
    
    \end{tabular}
    \caption{Performance of UniMatch V2 in the scenario of remote sensing change detection. We evaluate it on two representative datasets \textbf{LEVIR-CD}~\cite{chen2020spatial} and \textbf{WHU-CD}~\cite{ji2018fully}. The IoU$^c$ denotes \emph{changed-class IoU}, and OA denotes \emph{overall accuracy}.}
    \label{tab:remote_sensing}
\end{table*}

\subsection{Towards A More Real-World SSS Setting}

We hope to highlight that, existing SSS works \emph{all} adopt an \emph{artificial} semi-supervised setting, where labeled images are selected from all available labeled data as a small subset, and manually treat the remaining ones as unlabeled images, \ie, not use their ground-truth labels. 

\emph{However, a more real-world and practical SSS setting should be:} we directly use \emph{all available} labeled images as our labeled set, and seek additional \emph{truly unlabeled} images as our unlabeled set. This setting is much more challenging than existing settings, since the competitive fully-supervised baseline is hard to further boost. But we believe it is necessary to make efforts towards this, since the ultimate goal of SSS is to benefit real-world applications. In most real-world scenarios, substantial images (\eg, 10K) have been annotated over time, but meantime a much larger pool of images (\eg, 100K) have not yet been manually labeled.

To this end, we evaluate our UniMatch V2 under three challenging but practical settings: (1) we use COCO all 118K labeled images and its officially provided 123K unlabeled images, (2) we use ADE20K all 20K labeled images, and treat the COCO 118K labeled images as our unlabeled data, and (3) we use Cityscapes all 3K labeled images, and treat its additionally provided 20K images without precise labels as our unlabeled pool. All the results are summarized in Table~\ref{tab:large_scale_sss}. There are four observations. (1) Even already given adequate labeled images, our UniMatch V2 can still enhance the model performance by incorporating additional unlabeled images. All the three challenging fully-supervised baselines are boosted by our UniMatch V2. (2) Even if the labeled and unlabeled images come from slightly different data distributions, unlabeled images are still highly beneficial. For instance, COCO 118K unlabeled images improve the ADE20K fully-supervised result (with 20K labeled images) from 54.1\% to 54.9\%. (3) There is a scaling law in unlabeled data. Notably, for the ADE20K dataset, when increasing the number of unlabeled COCO images from 118K to 241K, the performance is further improved from 54.9\% to 55.7\%. (4) Compared with small-scale SSS settings, our improvements in such real-world large-scale scenarios appear marginal, mainly due to the high labeled-only results. But we believe such a challenging direction is meaningful to explore and we expect future works can have significant advances.

\subsection{Extending UniMatch V2 to Broader Scenarios}

\subsubsection{Remote Sensing Change Detection}

Remote sensing images or satellite images are extremely laborious to annotate due to ultra-high resolution. Therefore, we further investigate the effectiveness of our UniMatch V2 in the task of remote sensing change detection, which aims to spot the changed regions between two bi-temporal images. A widely used pipeline is illustrated in Figure~\ref{fig:cd_pipeline}. Extracted features of the dual images are subtracted and then sent into the decoder. It can be considered a binary segmentation task, but with dual inputs. In this task, we use the student itself to produce pseudo labels, because it evolves much quicker than its EMA version. We train the model for 60 epochs, with batch size 8+8 and image size 252. Other hyper-parameters (\eg, learning rate, optimizer) are the same as those for natural images.

As shown in Table~\ref{tab:remote_sensing}, on the two most representative change detection datasets LEVIR-CD~\cite{chen2020spatial} and WHU-CD~\cite{ji2018fully}, our UniMatch V2, even building on the lightest DINOv2-S encoder, consistently surpasses all prior works across all settings. When adopting the larger DINOv2-B encoder, the results are further enhanced. For instance, on the WHU-CD dataset with 20\% labeled images, our DINOv2-S IoU result is superior to the previous best SemiCD-VL~\cite{li2024diffmatch} by 1.3\% (84.8\% \emph{vs.} 86.1\%), while our DINOv2-B result is better than it by 3.1\% (84.8\% \emph{vs.} 87.9\%). These impressive results demonstrate the strong universality of our framework, even for bi-temporal SSS scenarios.

\begin{table}[t]
\setlength\tabcolsep{1.28mm}
    \centering
    \begin{tabular}{lcccccc}
    \toprule

    \multirow{2}{*}{\textbf{CIFAR-10 Classification}} & \multicolumn{5}{c}{Seed} & \multirow{2}{*}{\textbf{Mean}} \\

    \cmidrule(lr){2-6}

    & 0 & 1 & 2 & 3 & 4 & \\
    
    \midrule

    SimMatch~\cite{simmatch} & 95.58 & 95.50 & 95.34 & 94.06 & 95.26 & 95.15 \\

    ShrinkMatch~\cite{shrinkmatch} & 95.39 & 95.44 & 95.36 & 94.76 & 95.35 & 95.26 \\
    
    \rowcolor{shadecolor}
    UniMatch V2 & 95.70 & 95.42 & 95.54 & 95.09 & 95.34 & \textbf{95.42} \\
    
    \bottomrule
    
    \end{tabular}
    \caption{Results of our UniMatch V2 of the semi-supervised \textbf{classification} task on \textbf{CIFAR-10} with 25 labels per class.}
    \label{tab:cifar10}
\end{table}

\begin{table}[t]
\setlength\tabcolsep{3.4mm}
    \centering
    \begin{tabular}{lcccc}
    \toprule

    \multirow{2}{*}{\textbf{STL-10 Classification}} & \multicolumn{3}{c}{Seed} & \multirow{2}{*}{\textbf{Mean}} \\

    \cmidrule(lr){2-4}

    & 0 & 1 & 2 & \\
    
    \midrule

    FixMatch~\cite{fixmatch} & 90.91 & 88.71 & 90.94 & 90.19 \\

    FlexMatch~\cite{flexmatch} & 91.35 & 92.29 & 91.66 & 91.77 \\

    ShrinkMatch~\cite{shrinkmatch} & 91.13 & 92.43 & 91.10 & 91.55 \\
    
    \rowcolor{shadecolor}
    UniMatch V2 & 91.70 & 92.91 & 92.75 & \textbf{92.45} \\
    
    \bottomrule
    
    \end{tabular}
    \caption{Results of our UniMatch V2 of the semi-supervised \textbf{classification} task on \textbf{STL-10} with 25 labels per class.}
    \label{tab:stl10}
\end{table}

\begin{table}[t]
\setlength\tabcolsep{1.2mm}
    \centering
    \begin{tabular}{lcccccccc}
    \toprule
    
    \multirow{2}{*}{\textbf{SVHN}} & \multicolumn{3}{c}{Seed (4 labels)} & \multirow{2}{*}{\textbf{Mean}} & \multicolumn{3}{c}{Seed (25 labels)} & \multirow{2}{*}{\textbf{Mean}} \\
    
    \cmidrule(lr){2-4}\cmidrule(lr){6-8}

    & 0 & 1 & 2 & & 0 & 1 & 2 & \\
    
    \midrule

    FixMatch~\cite{fixmatch} & 94.5 & 96.9 & 97.1 & 96.2 & 98.0 & 98.0 & 97.9 & 98.0 \\

    FlexMatch~\cite{flexmatch} & 89.2 & 89.9 & 96.3 & 91.8 & 91.8 & 91.8 & 96.7 & 93.4 \\

    ShrinkMatch~\cite{shrinkmatch} & 98.0 & 97.8 & 96.7 & 97.5 & 98.1 & 98.1 & 98.0 & 98.0 \\
    
    \rowcolor{shadecolor}
    UniMatch V2 & 98.2 & 98.2 & 98.1 & \textbf{98.2} & 98.2 & 98.2 & 98.2 & \textbf{98.2} \\
    
    \bottomrule
    
    \end{tabular}
    \caption{Results of our UniMatch V2 of the semi-supervised \textbf{classification} task on \textbf{SVHN} with 4 or 25 labels per class.}
    \label{tab:svhn}
\end{table}

\begin{table}[t]
\setlength\tabcolsep{1.5mm}
    \centering
    \begin{tabular}{lccg}

    \toprule
    
    \textbf{ImageNet-1K} & CoMatch~\cite{li2021comatch} & SimMatch~\cite{simmatch} & UniMatch V2 \\

    \midrule
    
    Top-1 / Top-5 Acc & 73.6 / 91.6 & 74.1 / 91.5 & \textbf{74.3} / \textbf{91.7} \\

    \bottomrule
    
    \end{tabular}
    \caption{Results of our UniMatch V2 of the semi-supervised \textbf{classification} task on \textbf{ImageNet-1K} with 10\% labels.}
    \label{tab:imagenet}
\end{table}

\begin{figure}[t]
    \centering
    \includegraphics[width=\linewidth]{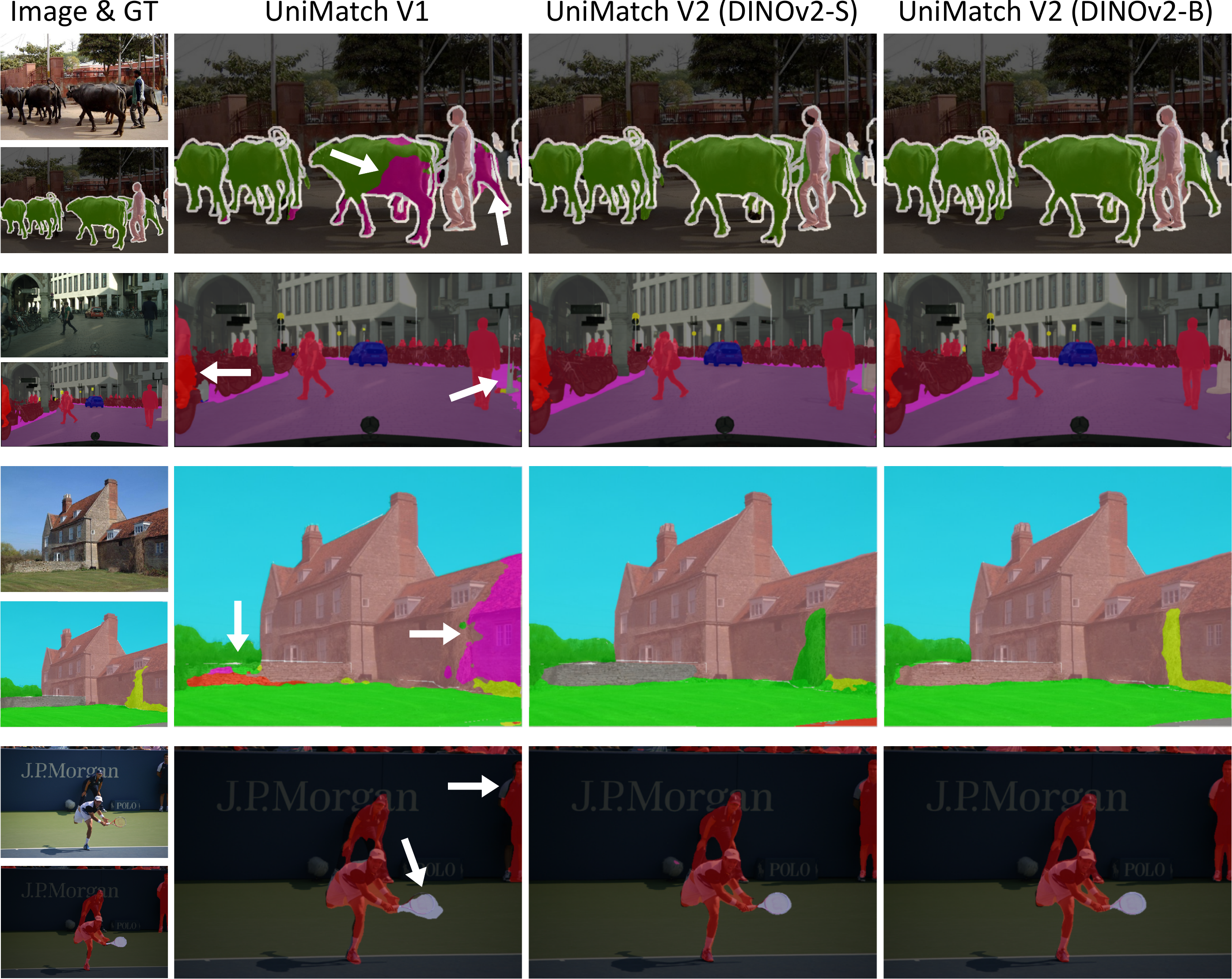}
    \caption{Qualitative comparisons between UniMatch V1~\cite{unimatch} and current V2. From top to bottom, the images are sampled from Pascal, Cityscapes, ADE20K, and COCO datasets, respectively.}
    \label{fig:vis_comparison_v1}
\end{figure}

\begin{figure}[t]
    \centering
    \includegraphics[width=\linewidth]{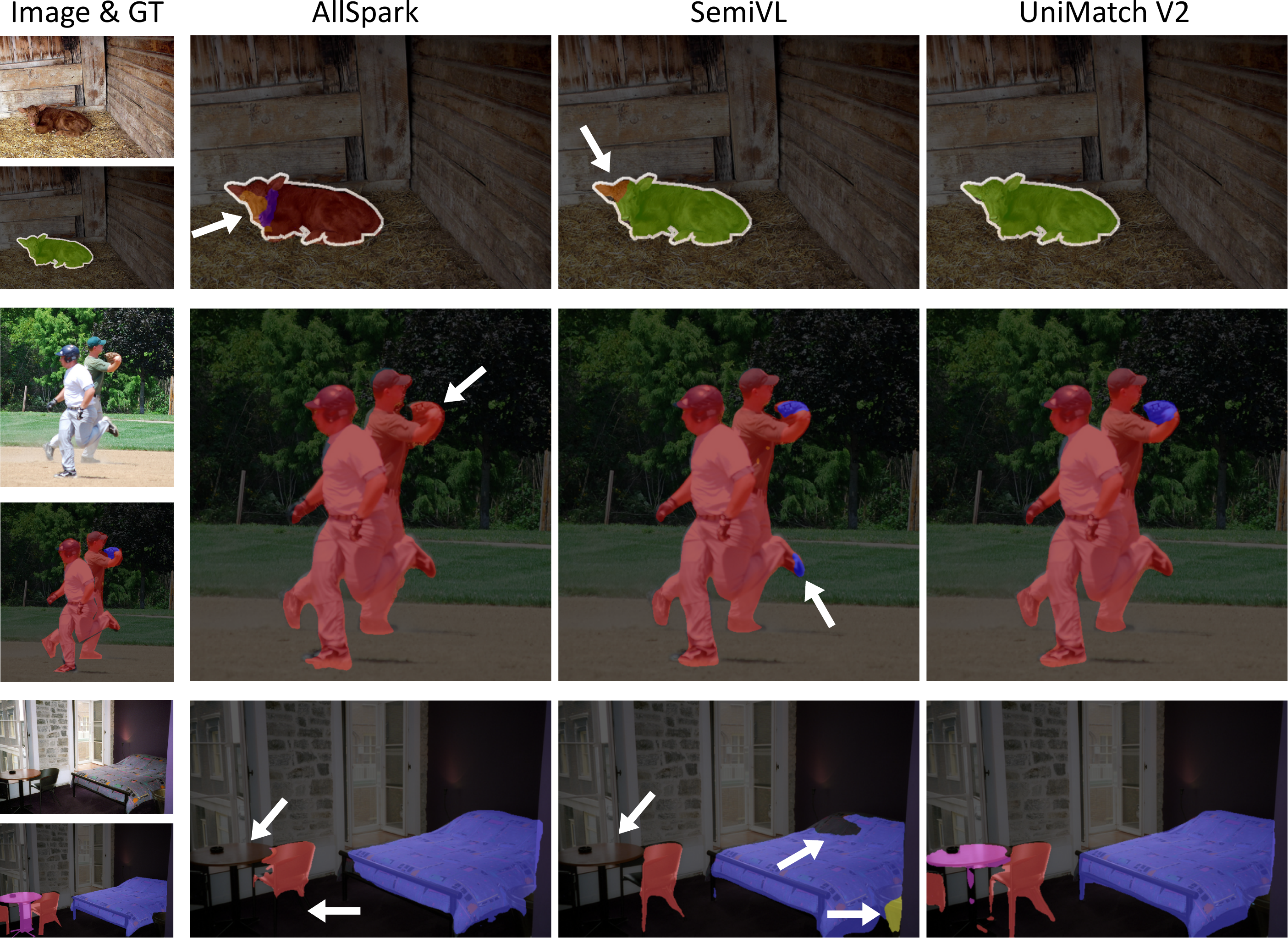}
    \caption{Qualitative comparisons with previous state-of-the-art methods AllSpark~\cite{allspark} and SemiVL~\cite{semivl}.}
    \label{fig:vis_comparison_others}
\end{figure}

\subsubsection{Semi-Supervised Classification}

We further extend our UniMatch V2 framework to the more fundamental scenario of semi-supervised classification. We follow the protocols of previous works~\cite{fixmatch, simmatch} and conduct experiments on four representative datasets. On CIFAR-10~\cite{cifar} (Table~\ref{tab:cifar10}), STL-10~\cite{stl10} (Table~\ref{tab:stl10}), and SVHN~\cite{svhn} (Table~\ref{tab:svhn}), we adopt FixMatch~\cite{fixmatch} with distribution alignment~\cite{remixmatch} as our baseline. For stability of reported performance, we run each labeled split with three or five different random seeds. On ImageNet-1K~\cite{deng2009imagenet} (Table~\ref{tab:imagenet}), we use SimMatch~\cite{simmatch} as our baseline. As reported in the tables, on all the four datasets, our UniMatch V2 outperforms previous methods non-trivially.

\subsection{Qualitative Comparisons}

We qualitatively compare our current UniMatch V2 framework with our previous UniMatch V1 in Figure~\ref{fig:vis_comparison_v1}. On the four representative benchmarks, our UniMatch V2 produces sharper predictions than V1, also with less confusion to challenging semantics. Additionally, we provide qualitative comparisons with previous SOTA methods AllSpark~\cite{allspark} and SemiVL~\cite{semivl} in Figure~\ref{fig:vis_comparison_others}. Our UniMatch V2 stands out as the most accurate semantic segmentor.
\section{Conclusion}

In this work, we present UniMatch V2 to strengthen our prior V1 framework for semi-supervised semantic segmentation. We update previous outdated ResNet encoders to the most capable DINOv2 encoders. We conduct comprehensive experiments for future works to compare with us in this new benchmark easily. Technically, we unify the image-level and feature-level augmentations into a single stream, and further design a Complementary Dropout to take full advantage of the dual-stream practice by crafting better dual learnable views. Consequently, our UniMatch V2 significantly outperforms all precedent works.

\bibliographystyle{IEEEtran}
\bibliography{egbib}

\end{document}